\documentclass[11pt]{article}

\usepackage{acl}

\usepackage{times}
\usepackage{latexsym}
\usepackage{booktabs}   
\usepackage{url}
\usepackage{longtable}    
\usepackage{todonotes}
\usepackage{pdflscape}  
\usepackage{algorithm}
\usepackage{algpseudocode}
\usepackage{amsmath}
\usepackage{amssymb}
\usepackage{tabularx}
\usepackage{booktabs}
\usepackage{float}
\usepackage[normalem]{ulem}
\newcommand\blfootnote[1]{%
  \begingroup
  \renewcommand\thefootnote{}\footnote{#1}%
  \addtocounter{footnote}{-1}%
  \endgroup
}
\usepackage[T1]{fontenc}

\usepackage[utf8]{inputenc}

\usepackage{microtype}

\usepackage{inconsolata}

\usepackage{graphicx}

\title{A Pre-Training Analogue of Grokking in Language Models: Tracing Delayed Grammatical Generalization}
\author{Sherin Muckatira, Namrata Shivagunde, Vijeta Deshpande, Anna Rumshisky \\ 
University of Massachusetts Lowell\\ 
\texttt{sherinbojappa\_muckatira@student.uml.edu} 
 }

\begin{document}
\maketitle
\begin{abstract}
Grokking, the phenomenon in which neural networks generalize long after
fitting their training data, has been studied in supervised settings on many epochs.  LLM pre-training instead involves next-token prediction over an unlabeled corpus, with limited data repetition and no explicit train/validation split. To address this, we propose an exposure-based framework that enables the study of grokking-like dynamics during LLM pre-training. We ground our evaluation in BLiMP minimal pairs, which provide controlled grammatical contrasts. For every BLiMP minimal pair, we identify a critical phrase, the smallest continuous span that captures the grammatical contrast and the phenomenon-relevant context. Examples whose critical phrase appears in the pre-training window are assigned to the proxy-train split; the remaining examples are assigned to the proxy-validation split. Across five grammatical phenomena, we observe delayed generalization. Analyzing pre-training checkpoints before and after generalization shows that most of the code length compression in the proxy-validation split happens within the delayed generalization window, grammatical concept vectors become more predictive of grammatical acceptability and occupy a higher-dimensional subspace after generalization. We also find that attention from the critical token to the relevant context token is concentrated in a small number of heads.
\blfootnote{\raggedright Code is available at \url{https://github.com/SherinBojappa/pre_training-analogue-of-grokking}}
\end{abstract}

\begin{figure*}[t]
    \centering
    \includegraphics[width=\textwidth]{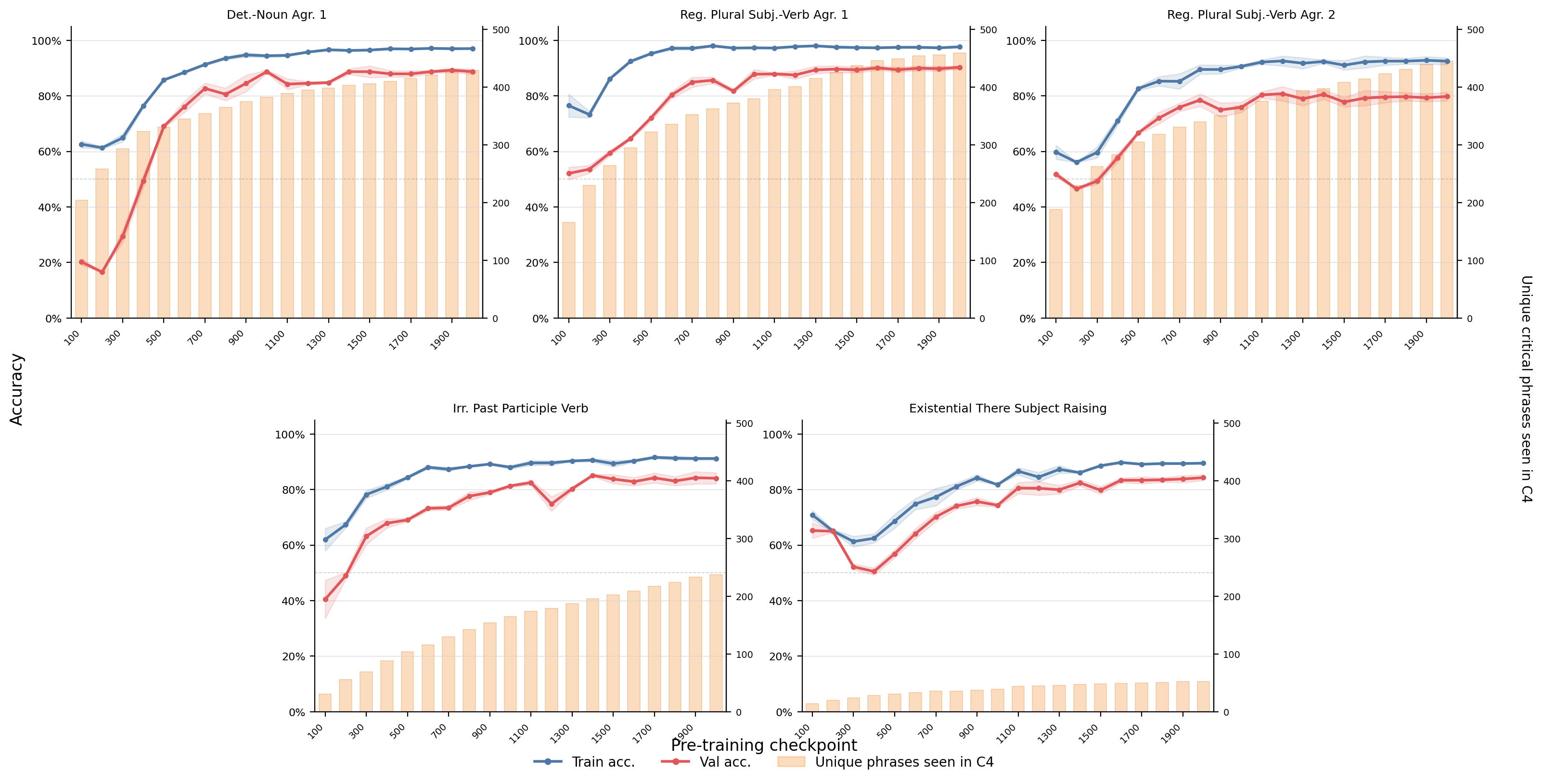}
    \caption{Delayed-generalization curves and cumulative C4 exposure for the analyzed BLiMP phenomena. Blue and red lines show mean accuracy on the proxy-train (exposed), and proxy-validation (unexposed) splits across three random seeds; x-axis shows training steps; shaded bands indicate $\pm 1$ standard error. Orange bars show the cumulative number of unique critical phrases encountered in the C4 pre-training stream up to each checkpoint. Across phenomena, proxy-train accuracy rises before proxy-validation accuracy.}
    \label{fig:comb_grokking}
\end{figure*}

\section{Introduction}



Grokking is a phenomenon in neural networks where models transition from fitting training examples to generalizing to held-out examples. In the classical setting, models are trained for many epochs on a fixed supervised train split and evaluated on a held-out validation split drawn from the same task distribution \citep{power2022grokkinggeneralizationoverfittingsmall}. 
This delayed transition provides
a well-defined behavioral boundary for analyzing internal changes associated with the onset of generalization.

A pre-training formulation of grokking would provide a way to identify when an ability begins to generalize during training. 
Training loss alone does not reveal which abilities are acquired at a given stage. This is relevant because frontier models increasingly use multi-stage pre-training, with changes in data composition \citep{yang2025qwen3technicalreport}. Skill-specific generalization transitions could therefore inform when to change the data mixture based on whether the targeted abilities have begun to generalize.

However, classical grokking is defined with respect to a particular task:
models are trained on a subset of task examples and delayed generalization is
measured on held-out examples from the same task. This setting does not directly carry over to language-model pre-training. During pre-training, the model is optimized for next-token prediction over a large unlabeled corpus, rather than trained directly on labeled examples from
the downstream evaluation task. As a result, there is no natural supervised
train/validation split over downstream examples. To bring the classical grokking formulation into the pre-training setting, we construct an exposure-based proxy split over downstream evaluation datasets. 

We address this problem by constructing an exposure-based proxy split for
studying delayed generalization during language-model pre-training. We use
grammar as a test case because grammatical dependencies are structured,
measurable, and can be tested using controlled minimal-pair datasets.
In
particular, we use BLiMP \citep{warstadt-etal-2020-blimp-benchmark}, where each example contains an acceptable sentence and a minimally different unacceptable
sentence targeting a specific grammatical phenomenon. For each minimal pair, we
extract a critical phrase: a contiguous span that contains the grammatical
difference and the context needed to evaluate the grammatical dependency. We then assign
examples to either a proxy-train split if the critical phrases appear verbatim in
the C4 \citep{raffel2020exploring} tokens seen during the analyzed pre-training window, or to a
proxy-validation split otherwise.

This split does not imply that the model has never seen the relevant
grammatical rule. It only distinguishes examples with verbatim
critical-phrase exposure from examples without such exposure. This gives a
pre-training analogue of the classical grokking setup: the model may first
perform well on examples with direct phrase exposure and only later improve
on examples without direct exposure.

Using this framework, we study early pre-training checkpoints of
35M, 60M, and 130M parameter LLaMA-style language models on five BLiMP
grammatical phenomena. We first measure whether examples with verbatim
critical-phrase exposure reach high accuracy before examples without such
exposure, and test this effect against matched random splits that preserve
split sizes. We then analyze grammar-specific contrast representations and
attention patterns before and after the delayed-generalization transition to
ask whether the behavioral lag is accompanied by changes in the model
representations of grammatical concepts and attention patterns. We find that proxy-train accuracy consistently reaches high accuracy
before proxy-validation accuracy, revealing delayed generalization. Checkpoint-level analyses further show
that this transition is accompanied by changes in grammar-specific concept
vectors and attention-to-context patterns.

Our contributions are:
\begin{enumerate}
    \item We introduce an exposure-based framework for studying
    grokking-like delayed generalization during language-model pre-training.
    \item We show that, across five grammatical phenomena, examples with
    verbatim critical-phrase exposure reach high accuracy before examples
    without such exposure.
    \item We validate this effect with matched random-split permutation
    tests, showing that the observed lag is not explained by split size
    alone.
    \item We show that the transition is accompanied by changes in
    grammar-specific concept vectors and attention-to-context patterns.
\end{enumerate}

\section{Related Work}

\paragraph{Grokking and delayed generalization.}
Grokking was introduced as a delayed transition from memorization to generalization, where models first reach high training accuracy while validation accuracy remains low, and only later generalize to held-out
examples \citep{power2022grokkinggeneralizationoverfittingsmall}. Most prior work studies this behavior in controlled settings with explicit train/validation splits, such as modular arithmetic and other algorithmic tasks \citep{gromov2023grokkingmodulararithmetic,
nanda2023progress}. Mechanistic analyses suggest that the behavioral transition can be preceded by internal changes in circuits or representations
\citep{nanda2023progress}. Subsequent work extends grokking to richer
language-like synthetic tasks involving hierarchical structure
\citep{murty-etal-2023-grokking}. We study a pre-training setting in
which the model is trained on next-token prediction over natural language,
rather than directly on a supervised downstream task.

\paragraph{Grokking-like dynamics during language-model pre-training.}
Recent work has begun to ask whether grokking-like dynamics occur during language-model pre-training. \citet{li2026grokking} analyze grokking in large mixture-of-experts models. However, they evaluate on downstream tasks after applying LoRA fine-tuning, making it unclear whether the observed dynamics are due to pre-training alone. They also define grokking solely via training loss, without a validation split. \citet{lv-etal-2025-language} argue that language models develop copying ability in a grokking-like manner, with induction heads emerging during pre-training. These works move grokking beyond synthetic tasks, but they do not directly adapt the classical train/validation formulation to downstream evaluation examples. We address
this gap by constructing an exposure-based proxy split: downstream evaluation examples are separated according to whether their critical phrases appear verbatim in the pre-training corpus. This lets us measure whether performance improves first on examples with direct corpus exposure and only later on examples without such exposure.

\paragraph{Linguistic generalization and representation geometry.}
BLiMP provides controlled minimal pairs for evaluating grammatical
knowledge in language models. Prior work using BLiMP shows that different grammatical phenomena follow different learning trajectories across checkpoints \citep{bunzeck-zarriess-2024-fifty}. We use BLiMP for a different purpose: to test whether grammatical minimal pairs exhibit delayed generalization during pre-training.

\paragraph{Representational geometry during learning.}
Prior work suggests that generalization can be accompanied by changes in the geometry of model representations. Studies of transformer training dynamics find that hidden-state geometry changes over training, including shifts in anisotropy, intrinsic dimensionality, and overall representation complexity \citep{razzhigaev-etal-2024-shape, li2026tracing}. Rather than measuring the geometry of hidden-state representations across all tokens, we focus on grammar-specific concept vectors. This is motivated by work that treats concepts or behaviors as directions in activation space \citep{park2024the,rimsky-etal-2024-steering}. In our use of concept vectors, we do not aim to steer the model or to establish a general theory of linear representations. Instead, we use grammar-specific concept directions as a diagnostic
tool for tracking how the separability and dimensional structure of
grammatical concepts change across pre-training checkpoints.

\paragraph{Generalization and memorization in language models.}
 \citet{mccoy-etal-2023-much, merrill-etal-2024-evaluating} assess whether the outputs of a fully trained model are novel with respect to its pre-training data. \citet{Hupkes_2023} provides a broader taxonomy for understanding generalization. Our work differs from 
 \citet{Hupkes_2023, mccoy-etal-2023-much, ruzzetti-etal-2026-lexical} by studying when generalization emerges across pre-training checkpoints and how internal representations change during that transition. \citet{ruzzetti-etal-2026-lexical} is especially relevant because it asks if examples containing more frequent lexical items are easier for fully trained models. It finds that lexical popularity has only a small effect on grammatical tasks in fully trained models. This is consistent with our checkpoint-level findings: exposure to the target pattern is important early in pre-training, but the performance gap between exposed and unexposed examples narrows as training progresses. These works primarily characterize generalization through black-box evaluation of model behavior. In contrast, we study when generalization emerges across pre-training checkpoints and examine how internal representations change during that transition.

\section{Method}
\label{sec:method}
In this section, we describe the pre-training setup, the exposure-based proxy split used to measure delayed generalization, and checkpoint-level representation and attention analyses.

\subsection{Pre-training and Evaluation Setup}

We train LLaMA-style decoder-only language models \citep{touvron2023llamaopenefficientfoundation} with 35M, 60M, and 130M parameters from the codebase released by \citet{zhao2024galore}. Models are trained on English C4 \citep{raffel2020exploring} with context length 256 and batch size 512, for 2000 steps with checkpoints every 100 steps. The model is trained on approximately 200M non-padding tokens. We use a cosine learning-rate schedule. The 35M and 60M models use learning rate \(10^{-3}\), and the 130M model uses learning rate \(5\times 10^{-4}\). The 60M model is trained with three independent random
seeds and is used for the main checkpoint-level analyses; the 35M and 130M models are used as scale robustness checks.

At each checkpoint, we evaluate on BLiMP using the EleutherAI evaluation harness \citep{eval-harness}. Each BLiMP example is a minimal pair containing one
acceptable and one unacceptable sentence. The model is correct if it assigns a 
higher probability to the acceptable sentence, making the chance accuracy 50\%. We
focus on the first 2000 pre-training steps because we do not want to analyze the fully saturated regime, but rather compare model behavior before and after proxy-validation accuracy begins to improve. We train our models on one NVIDIA RTX ADA 6000 48 GB GPU.


\subsection{Exposure-Based Proxy Split}

For each BLiMP minimal pair, we extract a \emph{critical phrase}: a contiguous span containing the token that differs between the acceptable and unacceptable sentence, together with the context needed to evaluate the
grammatical dependency. We extract critical phrases from both the acceptable and unacceptable sentences, lowercase them, and strip trailing punctuation. Dataset-specific extraction rules and examples are provided in
Appendix~\ref{app:critical_phrase}.

We define exposure by exact critical-phrase overlap with the C4 tokens seen during the analyzed pre-training window. A BLiMP example is assigned to the
proxy-train, or exposed, split if either its acceptable or unacceptable critical phrase appears verbatim in C4 during this window. It is assigned to
the proxy-validation, or unexposed, split otherwise. This exposure-based splitting procedure captures only verbatim critical-phrase exposure; it does not imply that the model has never
seen the underlying grammatical rule or related constructions.

\subsection{Delayed-Generalization Measurement}

Let \(A_{\mathrm{train}}(t)\) and \(A_{\mathrm{val}}(t)\) denote accuracy on
the proxy-train and proxy-validation splits at checkpoint \(t\), and let
\(\tau\) be an accuracy threshold. We define
\[
\begin{aligned}
t_{\mathrm{train}}(\tau)
&=
\min \{t : A_{\mathrm{train}}(t) \geq \tau \}, \\
t_{\mathrm{val}}(\tau)
&=
\min \{t : A_{\mathrm{val}}(t) \geq \tau \}.
\end{aligned}
\]
The delayed-generalization lag is
\[
\Delta t(\tau)=t_{\mathrm{val}}(\tau)-t_{\mathrm{train}}(\tau).
\]
We use \(\tau=80\%\) as the primary threshold because BLiMP is binary and
80\% represents substantial above-chance performance while still being
reached within the analyzed checkpoint range. Since checkpoints are saved
every 100 steps, transition times are checkpoint-level estimates. For
representation and attention analyses, we set
\(t_{\mathrm{before}}=100\) and
\(t_{\mathrm{after}}=t_{\mathrm{val}}(80\%)\). For the 60M model, we compute
\(t_{\mathrm{after}}\) separately for each seed and use the median value for
interpretability analyses. Appendix Table~\ref{tab:dataset_summary} reports
split sizes and transition checkpoints across thresholds.

\subsection{Dataset Selection}
\label{sec:dataset_selection}

Studying delayed generalization requires a ``Goldilocks'' regime: the dataset
must be neither so easy that accuracy saturates in the first few checkpoints,
nor so hard that proxy-validation accuracy never reaches a meaningful
threshold. We therefore select BLiMP datasets where a transition can be
measured within our 2000-step checkpoint window.

We apply four filtering stages. We exclude datasets that generalize too early
(mean accuracy over steps 100--300 \(\geq 70\%\)) or never reach high accuracy
(peak accuracy \(<80\%\)); datasets for which the targeted grammatical contrast and context dependency cannot be captured by a sub-sentential critical phrase; datasets with fewer than 100 proxy-train
examples after constructing the train/val splits, and datasets whose proxy-validation
split never reaches 80\%. These criteria select datasets where delayed generalization is measurable, but they do not condition on the proxy-train
split improving before the proxy-validation split. This yields six eligible
datasets.

After constructing the candidate datasets, we use the matched random-split
test below as a lag-validation step. We retain datasets whose exposure-based
lag is significant at \(p<0.05\) in at least two seeds. This excludes
Determiner--Noun Agreement 2, which shows zero lag and no significant seeds.
The final analysis set contains five grammatical phenomena. Full selection
details are provided in Appendix~\ref{sec:dataset_selection}.

\begin{figure*}[t]
    \centering
    \includegraphics[width=0.9\textwidth]{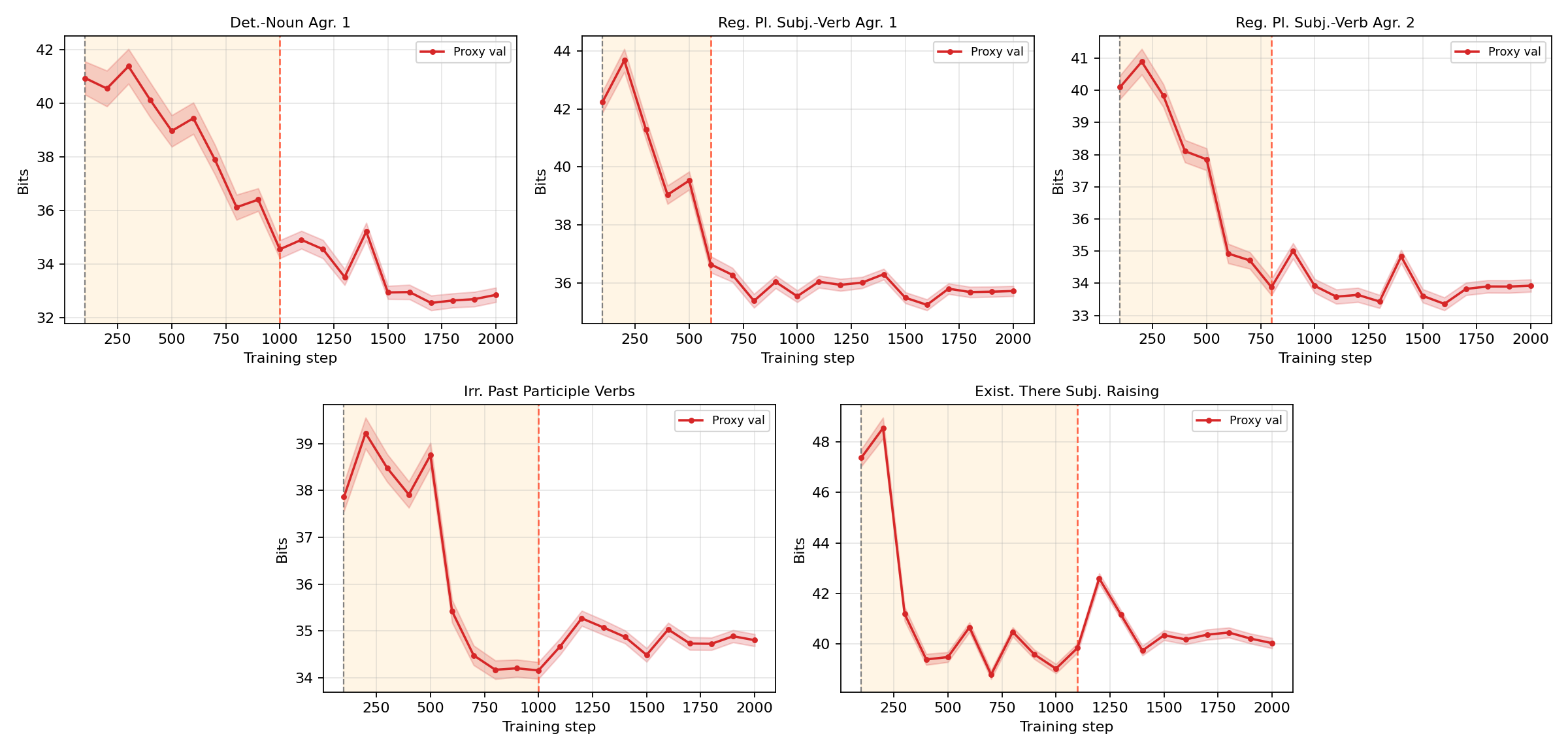}
    \caption{Code length of proxy-validation critical phrases across pre-training checkpoints. Shaded regions indicate the delayed-generalization window for each phenomenon. Code length decreases substantially during this interval for most phenomena and changes very little later on.}
    \label{fig:code_len_60M}
\end{figure*}

\subsection{Matched Random-Split Test}
\label{sec:random_split}
The observed lag between proxy-train and proxy-val
examples could arise from split size or arbitrary differences between
examples, rather than from corpus exposure. To test this, we compare the
exposure-based split to matched random splits. For each dataset and seed, we generate 1000 random splits that preserve the
number of proxy-train and proxy-val
 examples, but randomly reassign examples
independently of critical-phrase exposure. For each random split, we
recompute the checkpoint-level accuracy curves and measure the lag between the two groups reaching 80\% accuracy. This yields a null distribution of lags expected from arbitrary splits with the same group sizes. We then compare the delayed-generalization lag of the exposure-based proxy-train/proxy-validation split to this null distribution using a one-tailed empirical
\(p\)-value,

\[
p = \frac{c + 1}{n + 1},
\]

where \(c\) is the number of random splits whose lag is greater than or
equal to the observed exposure-based lag, and \(n=1000\) is the number of
random splits. Full seed-wise lags and $p$-values are reported 
in Appendix~\ref{app:permutation_test}.

\subsection{Code Length Measurement}
For each critical phrase corresponding to grammatical sentences in the validation set, we measure its code length in bits under the
checkpoint $\theta_t$:
\begin{equation}
    L(x_{1:n} \mid \theta_t)
    \;:=\;
    -\sum_{i=1}^{n}
    \log_2 p_{\theta_t}\!\bigl(x_i \mid x_0,\,x_{<i}\bigr),
    \label{eq:phrase-alone-bits}
\end{equation}
where $x_{1:n}$ is the critical phrase from the \emph{grammatical} sentence; $x_0$ is a single prefix token required to provide context for the first token of the phrase. It is the same constant for every phrase, split, and checkpoint;
$x_{<i} = (x_1,\dots,x_{i-1})$ is the left context tokens provided from the critical phrase for the subsequent tokens;
$\theta_t$ denotes the parameters of the checkpoint at
training step $t \in \{100, 200, \dots, 2000\}$.

\subsection{Grammatical Concept Vectors Analyses}
\label{sec:grammar_concept}
To analyze how grammatical information is represented across training, we
construct grammatical concept vectors from grammatical and ungrammatical BLiMP
pairs. For a given checkpoint and example \(i\), let
\(h^{(i)}_{\mathrm{gram}}\) and \(h^{(i)}_{\mathrm{ungram}}\) denote the
model hidden states for the acceptable and unacceptable sentence. We define the concept vector $v_i$

\[
v_i = h^{(i)}_{\mathrm{gram}} - h^{(i)}_{\mathrm{ungram}}.
\]

Unless otherwise stated, hidden states are taken from the final layer at
the last content-token position. Although these vectors are constructed from paired activation differences,
we refer to them as grammatical concept vectors because they represent the
grammatical concept expressed by each BLiMP minimal pair. This construction
is motivated by work treating concepts and behaviors as directions in
representation space \citep{park2024the, rimsky-etal-2024-steering}.

We analyze these vectors in two ways. First, we compute a mean-difference concept vector from the proxy-train split,

\[
\bar{v}_t =
\frac{1}{N}
\sum_{i=1}^{N}
\left(
h^{(i)}_{\mathrm{gram},t}
-
h^{(i)}_{\mathrm{ungram},t}
\right),
\]

and evaluate whether this direction separates grammatical from ungrammatical
sentences in the proxy-validation split. Similar to the mean-difference scoring approach of \citet{zou2025representationengineeringtopdownapproach}, we score each sentence in the proxy-validation split by its dot product
with \(\bar{v}_t\) and report AUROC.

Second, we measure the
effective rank of the mean-centered concept-vector matrix \citep{7098875}. For each checkpoint \(t\), we form a matrix \(V_t \in \mathbb{R}^{N \times d}\) where \(N\) is the number of minimal-pair examples and \(d\) is the hidden-state dimension. 
The \(i\)-th row of this concept-vector matrix is the example-level concept vector
\[
v_{i,t} = h^{(i)}_{\mathrm{gram},t} - h^{(i)}_{\mathrm{ungram},t}.
\]
We mean-center \(V_t\) across examples and compute its singular values. Given nonzero
singular values \(\sigma_1,\ldots,\sigma_r\), we normalize
\(\hat{\sigma}_k=\sigma_k/\sum_j\sigma_j\) and define
\[
\mathrm{\text{effective rank}}=\exp\left(-\sum_{k=1}^{r}\hat{\sigma}_k
\log \hat{\sigma}_k\right).
\]
A higher effective rank indicates that the grammatical concept vectors are distributed across
more independent directions.

\subsection{Attention-to-Context Analysis}
For each minimal pair, we identify the critical token
\(t_{\mathrm{critical}}\), whose form differs between the acceptable and
unacceptable sentence, and the earlier context token
\(t_{\mathrm{context}}\), which is needed to determine the correct
grammatical form. At each checkpoint, layer, and head, we extract the
attention weight from \(t_{\mathrm{critical}}\) to
\(t_{\mathrm{context}}\), averaged over all minimal pairs in the dataset. We
then measure the change in this attention-to-context score from
\(t_{\mathrm{before}}\) to \(t_{\mathrm{after}}\) and identify the heads with
the largest increase. We also measure the attention entropy at \(t_{\mathrm{critical}}\) to identify if attention becomes more spread out or concentrated after generalization. 

\section{Results}
We first show that the exposure-based proxy split reveals delayed
generalization, then analyze code length, concept-vector geometry, and attention dynamics across the transition.

\begin{figure*}[t]
    \centering
    \includegraphics[width=0.9\textwidth]{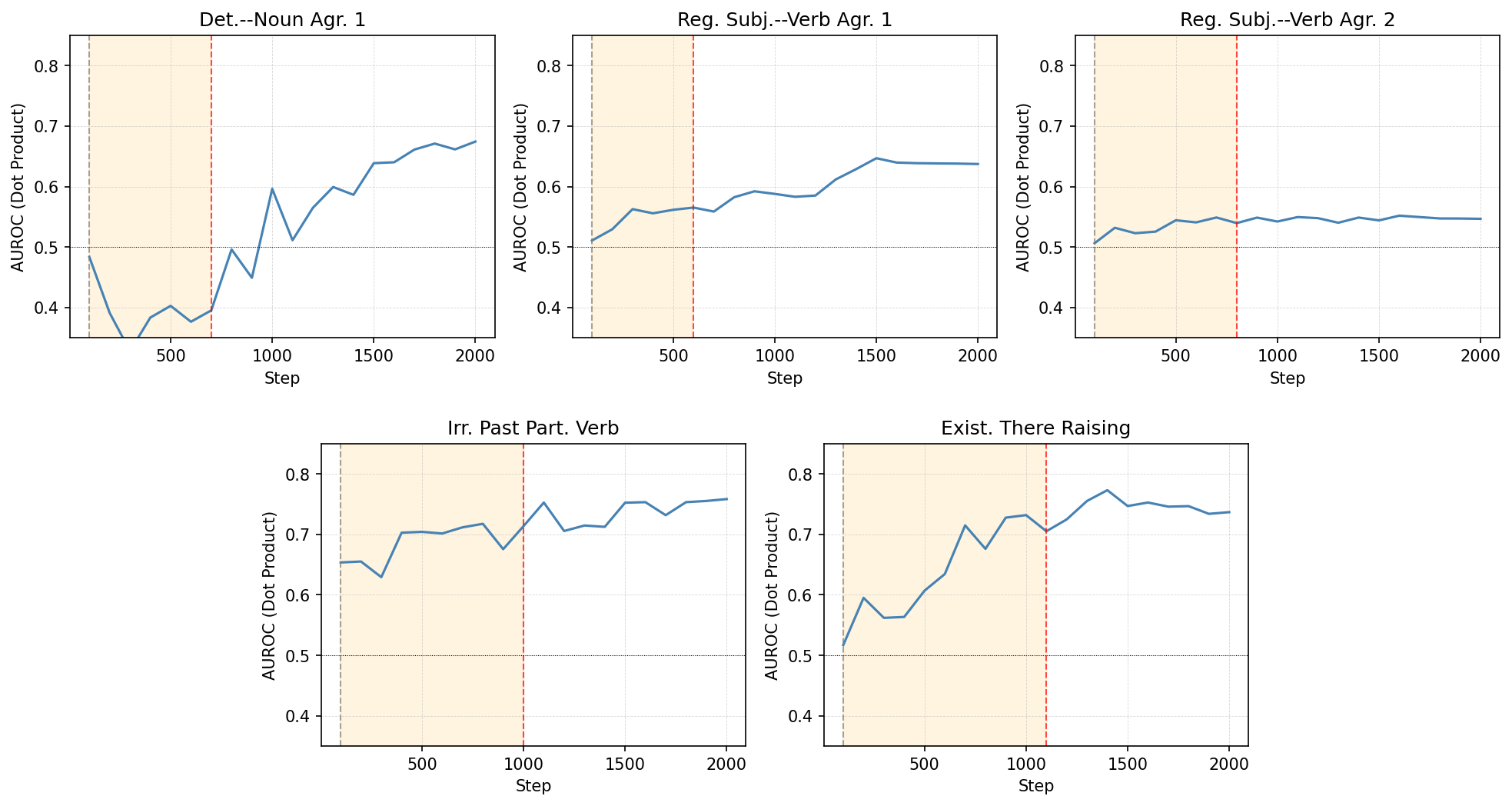}
    \caption{Concept-vector AUROC on the proxy-validation split across pre-training checkpoints. At each checkpoint, we compute a mean-difference vector from proxy-train minimal pairs and score proxy-validation sentences by their dot product with this vector. AUROC measures how well these scores separate grammatical from ungrammatical sentences using final-layer last-token representations.}
    \label{fig:concept_auroc}
\end{figure*}


\subsection{Exposure-Based Splits Reveal Delayed Generalization}
Figure~\ref{fig:comb_grokking} shows accuracy on the proxy-train (exposed) and proxy-validation (unexposed) splits across early pre-training checkpoints averaged over three independently trained 60M-parameter models. Across the five analyzed BLiMP phenomena, the proxy-train split reaches high accuracy earlier than the proxy-validation split. This produces a positive delayed-generalization lag: examples whose critical phrases appear verbatim in the C4 pre-training window reach the 80\% accuracy threshold before examples without verbatim critical-phrase exposure. The delayed-generalization lag varies by grammatical phenomenon. Some phenomena,
such as \emph{Regular Subject--Verb Agreement 1}, show relatively short lags, whereas \emph{Irregular Past Participle Verbs} and \emph{Existential There Subject Raising} show larger delays. We use the 80\% accuracy
threshold as the primary transition criterion. 

We also run the same exposure-conditioned evaluation for the 35M and 130M models. In both settings, the proxy-train split generally reaches high accuracy before the proxy-validation split. The delayed-generalization lag varies across datasets and model sizes. These results show that the delayed-generalization lag is not specific to the 60M model. Delayed-generalization curves for these model sizes are shown in Appendix~\ref{app:model_size}.

We validate the delayed-generalization lag using the matched random-split test described in Section~\ref{sec:random_split}; full seed-wise results are reported in
Appendix~\ref{app:permutation_test}.
Table \ref{tab:permutation_summary} shows that five of the six eligible datasets exhibit significant exposure-based delayed-generalization lags in at least two seeds. \emph{Determiner--Noun Agreement 2} is the only eligible dataset with zero observed lag in all three seeds and no significant permutation-test result; we therefore exclude it from the main analysis and report it as a non-lagging eligible case. Matched random
splits do not reproduce the observed lag, indicating that it is not explained by split size alone.

\begin{table}[t]
\centering
\setlength{\tabcolsep}{2pt}
\renewcommand{\arraystretch}{1.05}
\small
\begin{tabular}{lcc}
\toprule
Dataset & Observed lag range & Significant seeds \\
\midrule
Det.--Noun Agr.\ 1 & 100--500 & 2/3 \\
Det.--Noun Agr.\ 2 & 0 & 0/3 \\
Subj.--Verb Agr.\ 1 & 300--500 & 3/3 \\
Subj.--Verb Agr.\ 2 & 300--900 & 2/2 \\
Irr.\ Past Participle & 500--600 & 3/3 \\
Exist.\ There Raising & 400--500 & 3/3 \\
\bottomrule
\end{tabular}
\caption{Matched random-split permutation test results. Observed lag range is the range of delays, in training steps, between the proxy-train and proxy-validation splits reaching 80\% accuracy across seeds. Significant seeds are those where the exposure-based lag exceeds the matched random-split null distribution at $p < 0.05$. For \emph{Subject–Verb Agreement 2}, one seed is excluded because the proxy-validation split does not reach 80\% accuracy within the training window.}
\label{tab:permutation_summary}
\end{table}

  \begin{figure*}[t]                                                          
      \centering                                        
      \includegraphics[width=0.9\textwidth]{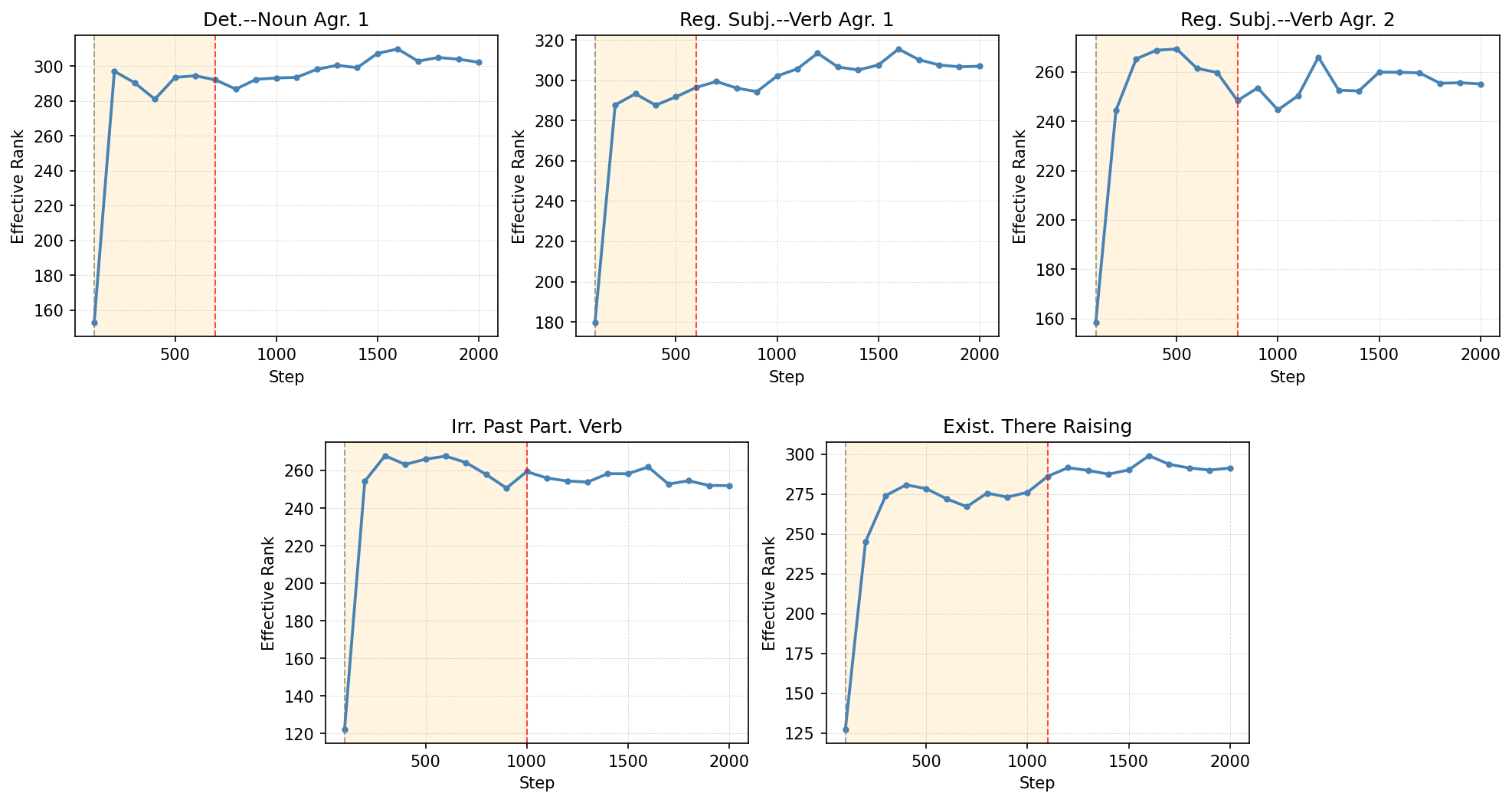}                               \caption{Effective rank of grammatical concept vectors across pre-training checkpoints for the five analyzed BLiMP phenomena. Higher effective rank indicates that the grammatical concept vectors span more independent directions in representation space. Dashed vertical lines mark $t_{\text{before}}=100$ and the dataset-specific $t_{\text{after}}$; the shaded region marks the delayed-generalization window.}
      \label{fig:rank_concept}                         
  \end{figure*} 

\paragraph{Continuous metrics and relationship to training loss.}
Prior work has shown that apparent emergent behavior can arise from the use of discontinuous evaluation metrics \citep{schaeffer2023are}. We therefore examine a continuous measure of grammatical preference by computing the log-likelihood margin for each minimal pair,
\[
m = \log P(\text{grammatical}) - \log P(\text{ungrammatical}),
\]
where positive values indicate that the model assigns higher likelihood to the grammatical sentence. Across most phenomena, the margin generally increases over the same training interval as accuracy, indicating that the observed transitions are also reflected in a continuous metric. C4 training loss decreases smoothly, while different grammatical phenomena improve at different checkpoints. Thus, global training loss captures overall model improvement but does not identify when individual grammatical abilities begin to generalize. Results are reported in Appendix Table \ref{tab:margin-training-loss}.

\subsection{Most Code-Length Reduction Occurs During Delayed Generalization Window}

Inspired by \citet{morris2026how}, who use model-based compression to study memorization, we track the code lengths of grammatical critical phrases across pre-training checkpoints. In figure \ref{fig:code_len_60M} we find that the reduction in code length of the critical phrases occur in the delayed generalization window and upon further training there is not much reduction. Together with the increase in grammatical accuracy, this suggests that the model learns the underlying grammatical pattern during this window. The subsequent plateau suggests diminishing returns from
continued training for these abilities, motivating skill-specific
generalization transitions as a potential signal for when changes to the
training data mixture may be useful.

\begin{figure*}[t]
    \centering
    \includegraphics[width=0.9\textwidth]{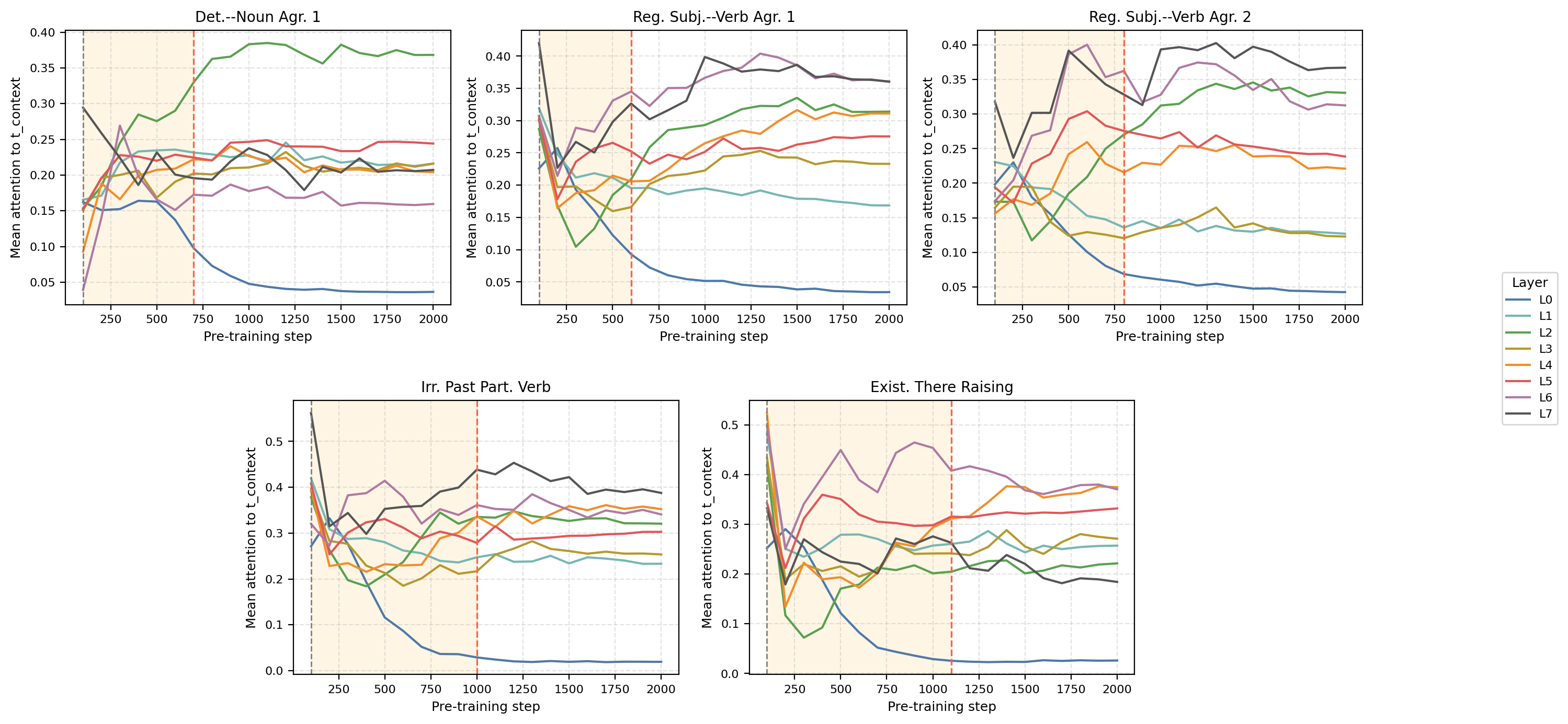}
    \caption{Average attention-to-context score across pre-training checkpoints. Each panel shows one BLiMP phenomenon; each line shows one layer, averaged over all 8 heads. The score is the attention weight from the critical token $t_{\text{critical}}$ to the earlier context token $t_{\text{context}}$ that determines the relevant grammatical dependency.}
    \label{fig:attn_context}
\end{figure*}

\subsection{Concept Vectors Capture Grammatical Separability}


Figure~\ref{fig:concept_auroc} tracks how well grammatical concept vectors
separate grammatical from ungrammatical sentences on the proxy-validation split
across pre-training checkpoints as defined in \ref{sec:grammar_concept}. Higher AUROC indicates that
grammatical and ungrammatical sentences are more separable along the concept-vector direction.

Across the five analyzed phenomena, separability generally strengthens during
pre-training, but the trajectory differs by phenomenon. \emph{Determiner--noun agreement},  \emph{Regular subject--verb agreement 1}, and  \emph{Existential-there subject raising} show the clearest increases. Irregular past participles are already separable early and remain well above
chance. \emph{Regular subject--verb agreement 2} remains closer to chance. Overall, these results show that separability along grammatical concept vectors generally improves during pre-training, but the strength and timing of this improvement vary across grammatical phenomena. Results across all layers are provided in  Figure \ref{fig:concept_auroc_all_layers} in the appendix.

\subsection{Concept Vectors Become More Distributed After Generalization}
\label{sec:concept_rank}
We next ask how grammar-specific concept vectors change in dimensionality during and after delayed generalization.
Figure~\ref{fig:rank_concept} shows that the effective rank of grammatical concept vectors generally increases during early pre-training. At
\(t_{\text{before}}\), the rank is lower, suggesting that the grammatical concept is
concentrated in a smaller number of dominant directions. During the delayed-generalization window, the rank fluctuates.  We use the dataset-specific $t_{\text{after}}$ values reported in Table~\ref{tab:t_after}. After \(t_{\text{after}}\), the effective rank increases and remains higher than at \(t_{\text{before}}\) across all five agreement datasets. This pattern suggests that after delayed generalization, the grammatical concept vector is represented across a higher-dimensional subspace. Prior work \citep{li2026tracing, razzhigaev-etal-2024-shape} has shown
that hidden-state representations can become more compressed after generalization. In our analysis,  we find that for these concept 
vectors, effective rank generally increases after delayed generalization, suggesting that the grammatical concept does not collapse into a few
dominant directions. Instead, the grammatical concept vectors become distributed across more directions. Results
across all layers are provided in
Figure ~\ref{fig:rank_concept_all_layers} in the appendix.

\subsection{Attention-to-Context Changes Around Delayed Generalization}



Figure~\ref{fig:attn_context} shows the average attention-to-context score across early pre-training checkpoints. Rather than showing a uniform increase across all layers and heads, the attention changes are localized: some layers show clear increases around the delayed-generalization transition, while
others remain stable or decrease. The largest attention-to-context changes are concentrated in a small number of heads, and the strongest heads differ
across phenomena, as shown in Table~\ref{tab:best_heads}. The full
layer--head results are provided in Appendix Table~\ref{tab:delta_attn}.

\emph{Determiner--noun agreement} shows the strongest increase in an early layer, whereas the \emph{Subject--verb} phenomena show their strongest increases in later layers. Appendix Figure~\ref{fig:attn_entropy} shows that attention entropy also fluctuates early and then often stabilizes or decreases after \(t_{\mathrm{after}}\). Representative token-level heatmaps in Appendix
Figure~\ref{fig:token_attn} show a qualitative pattern: after
\(t_{\mathrm{after}}\), selected heads attend more strongly to
dependency-relevant context tokens. Together, these results suggest that delayed grammatical generalization is accompanied by a localized and phenomenon-specific reorganization of attention, rather than a uniform shift toward more concentrated attention.

\section{Conclusion}

We studied whether grokking-like delayed generalization can be observed during standard language-model pre-training. Because pre-training lacks an explicit supervised train/validation split, we construct an exposure-based proxy split over BLiMP minimal pairs. Using this split, we find a pre-training analogue of grokking. Unlike the prolonged many-epoch train--validation gap often studied in classical supervised grokking, our setting reveals a checkpoint-level lag between exposed and unexposed evaluation examples. Across five grammatical phenomena, proxy-train split reaches high accuracy before proxy-validation accuracy,
and matched random-split tests suggest that this lag is not explained by the split size alone. We further find that much of the reduction in code length of grammatical critical phrases occurs during the delayed-generalization window, with relatively little additional reduction under continued training.

We show that this behavioral transition is accompanied by changes in grammar-specific concept vector representations and attention patterns.
Grammatical and ungrammatical examples become increasingly separable,
concept vectors become distributed across higher-dimensional subspace after generalization, attention analyses show
increased attention from the critical token to the context token needed to resolve the grammatical dependency, with the largest changes concentrated in
different heads across phenomena. Together, these results
suggest that corpus exposure can be used to identify when specific abilities
begin to generalize during pre-training. This may be particularly useful for
frontier models trained with multi-stage pre-training and changing data
mixtures, where detecting skill-specific
generalization transitions could help determine when additional training on a
given data mixture yields diminishing returns and when the training distribution
might be changed.

\section{Ethical considerations}
This work uses publicly available datasets: C4 for pre-training and BLiMP for
evaluation. We use these resources for research and evaluation under their respective terms of use. We do not redistribute matched C4 documents or extracted text. The study is focused on understanding language-model training dynamics, and we
do not foresee direct negative societal impacts. 

\section*{Limitations}
Our framework relies on exact critical-phrase overlap as a proxy for exposure.
It does not mean that the model has never seen the same grammatical rule in other sentences. Thus, the proxy validation split should be interpreted as examples without verbatim critical-phrase exposure, not as examples with entirely unseen grammar.

This work tests the framework on grammatical phenomena, where BLiMP
minimal pairs make it possible to define a sub-sentential critical phrase. Future work could extend this framework beyond grammar to other settings where
the relevant behavior can be linked to a sub-sentential phrase or span.

\bibliography{custom}

\appendix
\section{Appendix}
\label{sec:appendix}
\subsection{Critical Phrase Extraction}
\label{app:critical_phrase}
For each BLiMP minimal pair, we extract a \emph{critical phrase}: the
contiguous span that contains the token differing between the grammatical and ungrammatical sentences, together with the context needed to evaluate the grammatical dependency. The critical phrase extraction rule is defined separately for each grammatical phenomenon. For Determiner--Noun Agreement 1, we extract the span from the determiner to the changed noun. For Regular Plural Subject--Verb Agreement 1, we
extract the span from the subject head noun to the changed verb. For Regular Plural Subject--Verb Agreement 2, we extract the span from the
changed subject noun to the following verb. For Irregular Past Participle
Verbs, the span is from the subject head noun to the changed verb form. For Existential There Subject Raising, we extract
the sentence-initial existential construction through the predicate that
determines whether the raising construction is licensed. Examples of critical phrases are shown in Table \ref{tab:blimp_examples} in the appendix. For every minimal pair, we extract critical phrases from both the acceptable and unacceptable sentences. Before matching against C4, we lowercase the phrases and strip trailing punctuation.

\begin{table*}[t]
\centering
\small
\renewcommand{\arraystretch}{1.15}
\begin{tabular}{p{0.30\linewidth} p{0.22\linewidth} p{0.22\linewidth} p{0.22\linewidth}}
\hline
\textbf{dataset} & \textbf{acceptable} & \textbf{unacceptable} & \textbf{critical phrases}\\
\hline
\path{blimp_regular_plural_subject_verb_agreement_1} & Paula references Robert. & Paula reference Robert. & Paula reference; Paula references\\
\path{blimp_regular_plural_subject_verb_agreement_2} & The students perform. & The student perform. & student perform; student performs \\
\path{blimp_determiner_noun_agreement_1} & Raymond is selling this sketch. & Raymond is selling this sketches. & this sketch; this sketches \\
\path{blimp_existential_there_subject_raising} & There is soon to be a cat existing. & There is willing to be a cat existing. & There is soon to; There is willing to\\
\path{blimp_irregular_past_participle_verbs} & The mushroom went bad. & The mushroom gone bad. & The mushroom went; The mushroom gone\\

\hline
\end{tabular}
\caption{Example BLiMP minimal pairs from the datasets used in this study. Each row shows one dataset with an acceptable sentence and its minimally different unacceptable counterpart.}
\label{tab:blimp_examples}
\end{table*}

Table~\ref{tab:blimp_examples} shows representative BLiMP examples from the datasets used in this work. Each row contains an acceptable sentence and a similar unacceptable sentence. The difference between the two sentences is used to extract the critical phrase for the exposure-based proxy split.

\section{Candidate Dataset Selection}
\label{sec:dataset_selection}
BLiMP contains 67 minimal-pair datasets spanning a range of grammatical
phenomena. To study delayed generalization, we require datasets where a
transition from near-chance to high accuracy is observable within our
checkpoint range. We apply a multi-stage set of criteria to filter out the datasets to study.

\paragraph{Stage 1: Accuracy-based filtering.}
  We exclude datasets where the model already generalizes early in training (mean accuracy over steps 100--300 $\geq 70\%$; 19 datasets) and datasets where the model never reliably
  generalizes (peak accuracy $< 80\%$; 27 datasets). Since BLiMP is a binary task (chance = 50\%), 80\% corresponds to 60\% of the
maximum above-chance performance. The accuracy criteria excludes 46 datasets.  The
  remaining 21 candidate datasets are listed in Table~\ref{tab:candidates}. 

  \paragraph{Stage 2: Phrase extractability.}
  Our proxy-split methodology requires extracting a sub-sentential critical phrase whose presence in the C4 training corpus determines whether an example is assigned to the train or
  validation split. For 6 of the 21 candidates, the grammatical dependency spans the full sentence, making sub-sentential extraction inapplicable: \textit{anaphor\_number\_agreement},
  \textit{anaphor\_gender\_agreement}, \textit{principle\_A\_case\_2}, \textit{tough\_vs\_raising\_2}, \textit{coordinate\_structure\_constraint\_object\_extraction}, and
  \textit{wh\_island}. Since BLiMP sentences are synthetically generated and the benchmark was released after the C4 corpus used in the training of the models in the study was collected, full BLiMP sentences do not appear verbatim in
   C4, making full-sentence matching uninformative. This leaves 15 datasets with extractable sub-sentential phrases.

  \paragraph{Stage 3: Minimum train split size.}
  After searching C4 for critical phrase matches, we require at least 100 examples in the train split to ensure a reliable accuracy curve. This stage excludes 4 datasets.

  \paragraph{Stage 4: Validation convergence.}
  Finally, we require that the validation split accuracy reaches at least 80\% by the end of training. Datasets where the validation curve does not converge indicate that the model
  fails to generalize beyond its training exposure, and therefore are not suitable for our study on grokking. This removes four additional datasets.

  Table~\ref{tab:train_val_splits} reports the train and validation split sizes and peak validation accuracy for all datasets with extractable phrases, organized by which stage they are filtered out.

  \begin{table*}[t]
  \centering
  \small
  \begin{tabular}{lrrr}
  \toprule
  \textbf{Dataset} & \textbf{Train} & \textbf{Val} & \textbf{Peak Val} \\
  \midrule
  determiner\_noun\_agreement\_1 & 873 & 127 & 89.2\% \\
  determiner\_noun\_agreement\_2 & 883 & 117 & 92.6\% \\
  regular\_plural\_subject\_verb\_agreement\_1 & 383 & 617 & 90.3\% \\
  irregular\_past\_participle\_verbs & 234 & 766 & 85.1\% \\
  existential\_there\_subject\_raising & 254 & 681 & 84.3\% \\
  regular\_plural\_subject\_verb\_agreement\_2 & 476 & 524 & 80.8\% \\
  \midrule
  irregular\_plural\_subject\_verb\_agreement\_1 & 501 & 499 & 75.3\% \\
  irregular\_plural\_subject\_verb\_agreement\_2 & 634 & 366 & 75.4\% \\
  determiner\_noun\_agreement\_irregular\_1 & 778 & 222 & 53.6\% \\
  determiner\_noun\_agreement\_irregular\_2 & 837 & 163 & 79.8\% \\
  \midrule
  determiner\_noun\_agreement\_with\_adjective\_1 & 19 & 981 & --- \\
  determiner\_noun\_agreement\_with\_adj\_2 & 33 & 967 & --- \\
  determiner\_noun\_agreement\_with\_adj\_irregular\_2 & 43 & 957 & --- \\
  determiner\_noun\_agreement\_with\_adj\_irregular\_1 & 67 & 933 & --- \\
    only\_npi\_licensor\_present & 0 & 1000 & --- \\
  superlative\_quantifiers\_1 & 0 & 1000 & ---\\
  \bottomrule
  \end{tabular}
    \caption
  {Proxy train/validation split sizes and peak proxy-validation accuracy for datasets with extractable sub-sentential critical phrases. Peak Val is the maximum proxy-validation accuracy across checkpoints, averaged over three seeds where available. Datasets are grouped by selection outcome: those passing all stages (top), those excluded by Stage 4 for insufficient validation convergence (middle), and those excluded by Stage 3 for fewer than 100 proxy-train examples (bottom). For datasets excluded at Stage 3, Peak Val is not reported because the proxy-train split is too small to form a reliable delayed-generalization curve.}
  \label{tab:train_val_splits}
  \end{table*}

\begin{table*}[t]
\centering
\small
\begin{tabular}{lccc}
\toprule
\textbf{Dataset} & \textbf{Early} & \textbf{Peak} & \textbf{Final} \\
\midrule
only\_npi\_licensor\_present & 59.2\% & 99.6\% & 96.1\% \\
anaphor\_number\_agreement & 47.4\% & 98.2\% & 97.1\% \\
determiner\_noun\_agreement\_1 & 56.7\% & 96.0\% & 95.7\% \\
determiner\_noun\_agreement\_2 & 50.3\% & 95.1\% & 93.3\% \\
determiner\_noun\_agreement\_with\_adjective\_1 & 58.0\% & 94.5\% & 94.2\% \\
regular\_plural\_subject\_verb\_agreement\_1 & 64.6\% & 93.5\% & 92.5\% \\
superlative\_quantifiers\_1 & 38.1\% & 93.5\% & 64.0\% \\
principle\_A\_case\_2 & 55.8\% & 92.7\% & 91.0\% \\
determiner\_noun\_agreement\_with\_adj\_2 & 48.9\% & 92.2\% & 91.4\% \\
regular\_plural\_subject\_verb\_agreement\_2 & 52.1\% & 90.7\% & 88.6\% \\
determiner\_noun\_agreement\_irregular\_2 & 54.0\% & 89.1\% & 86.2\% \\
anaphor\_gender\_agreement & 32.7\% & 88.9\% & 88.5\% \\
irregular\_plural\_subject\_verb\_agreement\_2 & 57.7\% & 88.4\% & 87.4\% \\
determiner\_noun\_agreement\_with\_adj\_irregular\_2 & 54.6\% & 88.1\% & 86.2\% \\
irregular\_past\_participle\_verbs & 60.9\% & 87.1\% & 82.3\% \\
tough\_vs\_raising\_2 & 65.7\% & 85.5\% & 83.5\% \\
determiner\_noun\_agreement\_with\_adj\_irregular\_1 & 52.4\% & 84.3\% & 82.8\% \\
existential\_there\_subject\_raising & 57.4\% & 84.1\% & 83.6\% \\
irregular\_plural\_subject\_verb\_agreement\_1 & 53.3\% & 81.8\% & 80.7\% \\
coordinate\_structure\_constraint\_object\_extraction & 50.6\% & 81.7\% & 80.9\% \\
wh\_island & 29.0\% & 81.2\% & 79.1\% \\
\bottomrule
\end{tabular}
\caption{The 21 candidate BLiMP datasets that pass the initial accuracy-based filtering criteria: early accuracy below 70\% and peak accuracy at least 80\%. Early is the mean accuracy over steps 100--300; Peak is the maximum accuracy across all checkpoints; Final is the mean accuracy over steps 1600--2000.}
\label{tab:candidates}
\end{table*}

\subsection{Full permutation-test results}
\label{app:permutation_test}

We perform a permutation test to evaluate whether the observed lag between exposed and unexposed examples is specific to the exposure-based proxy split. For each dataset and random seed, we preserve the sizes of the exposed and unexposed splits, but randomly reassign examples to the two groups independently of corpus exposure. This produces a null distribution over lags that would be expected from arbitrary splits of the same size.

For each of 1000 random permutations, we recompute the checkpoint-level accuracy curves for the two permuted groups. We then measure grokking strength which is defined as the lag, in training steps, between the first checkpoint at which the train and validation values reach 80\% accuracy. The observed exposure-based lag is compared against this null distribution using a one-tailed empirical $p$-value,
\[
p = \frac{c+1}{n+1},
\]
where $c$ is the number of permutations whose lag is greater than or equal to the observed lag, and $n=1000$ is the number of permutations.

Table~\ref{tab:permutation_pvalues} reports the full seed-wise results. Five of six datasets show significant exposure-based lags in at least two seeds. Subject--Verb Agreement 1, Irregular Past Participle Verbs, and Existential There Subject Raising are significant in all three seeds. Determiner--Noun Agreement 1 is significant in two of three seeds. Subject--Verb Agreement 2 is significant in both seeds; the remaining seed is not evaluated because the unexposed split does not reach the 80\% threshold within the training window. Determiner--Noun Agreement 2 is the only dataset with no significant seeds. These results show that the observed lags are not generally reproduced by arbitrary splits with the same group sizes.

\subsection{Dataset Summary}
\label{app:dataset_summary}
Appendix Table~\ref{tab:dataset_summary} reports the proxy-train and
proxy-validation split sizes for each dataset, together with the
proxy-validation transition checkpoints across thresholds from 70\% to 90\% on one seed.

  \begin{table*}[t]
  \centering
  \setlength{\tabcolsep}{4pt}
  \renewcommand{\arraystretch}{1.1}
  \small
  \begin{tabular}{lcc|ccccc}
  \toprule
   & & & \multicolumn{5}{c}{$t_{\text{after}}$} \\
  \cmidrule(l){4-8}
  Dataset & Train & Val & 70\% & 75\% & 80\% & 85\% & 90\% \\
  \midrule
  Det.-Noun Agr.\ 1 & 873 & 127 & 500 & 700 & 1000 & 1000 & 1400 \\
  Reg.\ Plural Subj.-Verb Agr.\ 1 & 383 & 617 & 600 & 600 & 600 & 700 & 1300 \\
  Reg.\ Plural Subj.-Verb Agr.\ 2 & 476 & 524 & 700 & 700 & 800 & 1200 & N/A \\
  Irr.\ Past Participle Verb & 234 & 766 & 500 & 600 & 800 & 1400 & N/A \\
  Existential There Subject Raising & 254 & 681 & 800 & 1100 & 1400 & N/A & N/A \\
  \bottomrule
  \end{tabular}
    \caption{Proxy train/validation split sizes and validation transition steps $t_{\text{after}}$ across different accuracy thresholds. Each threshold column reports the first saved pre-training checkpoint, in training steps, at which proxy-validation accuracy reaches that threshold. N/A indicates that the threshold is not reached within the 2000-step training window. Checkpoints are saved every 100 steps.}
  \label{tab:dataset_summary}
  \end{table*}

\section{Effect of Scale on Delayed Generalization}
\label{app:model_size}

\begin{figure*}[t]
    \centering
    \includegraphics[width=\textwidth]{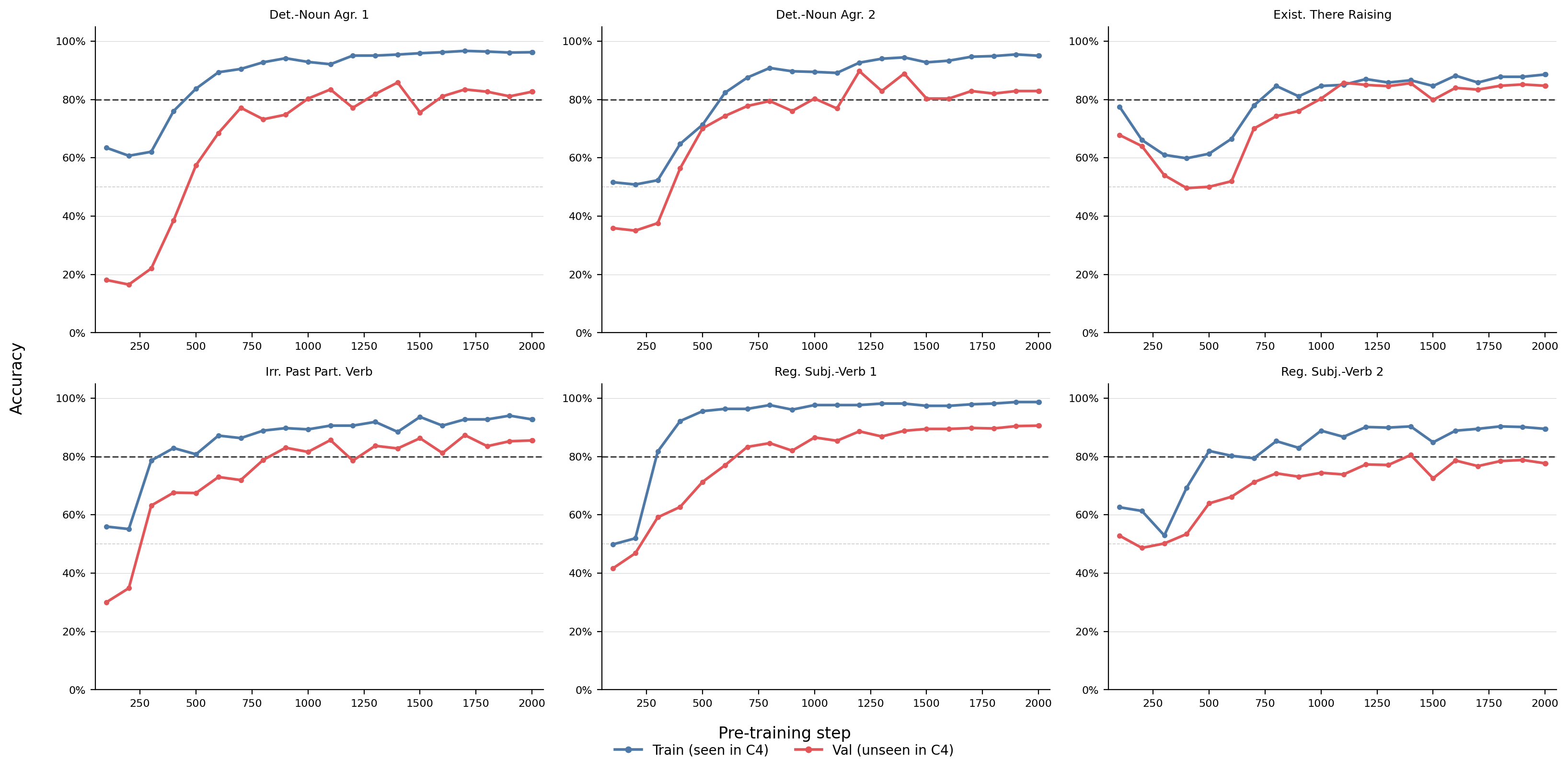}
    \caption{Delayed-generalization curves for the 35M  model. Blue and red lines show accuracy on the proxy-train and proxy-validation splits, respectively.}
    \label{fig:comb_grokking_35m}
\end{figure*}

\begin{figure*}[t]
    \centering
    \includegraphics[width=\textwidth]{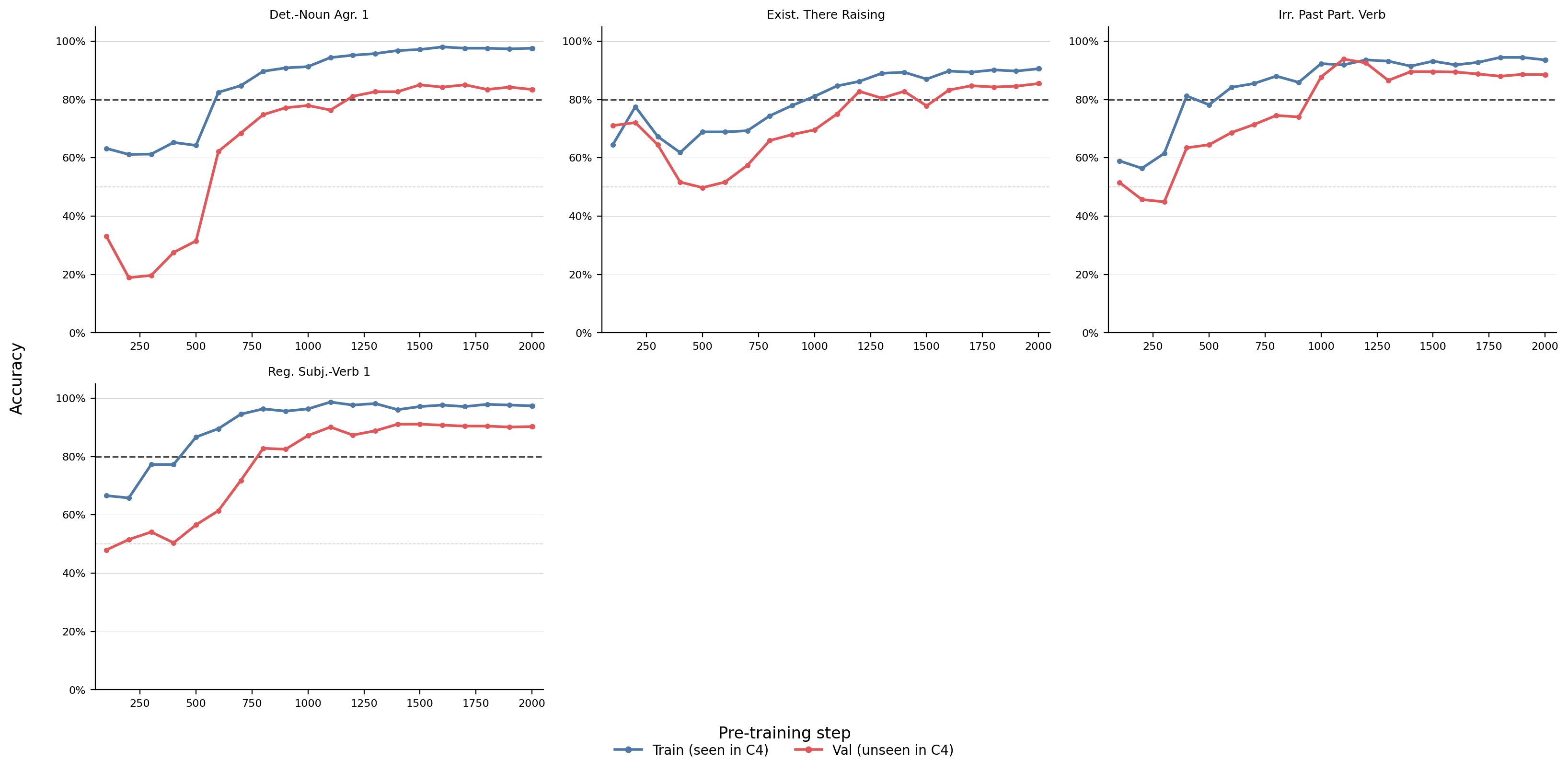}
    \caption{Delayed-generalization curves for the 130M model. Blue and red lines show accuracy on the proxy-train and proxy-validation splits, respectively.}
    \label{fig:comb_grokking_130m}
\end{figure*}
Figures~\ref{fig:comb_grokking_35m} and~\ref{fig:comb_grokking_130m} confirm
that grokking curves appear at both scales: training-split accuracy
consistently leads validation-split accuracy. However, the dynamics
differ across scales. The 35M model reaches the validation 80\%
threshold earlier than the 130M model on most datasets, despite being
the smaller model. 
This could be due to the fixed-budget capacity effect:
because all three models are trained for the same number of steps,
smaller models see more tokens per parameter and are more
capacity-constrained, which may force earlier extraction of
generalizable patterns. Larger models may require more training before
their additional capacity is converted into robust generalization.

Figures~\ref{fig:comb_grokking_35m} and~\ref{fig:comb_grokking_130m}
show that the delayed-generalization pattern is also visible at other
model scales. In both the 35M and 130M models, the proxy train split
generally reaches high accuracy before the proxy validation split, indicating
that the delayed-generalization pattern is not specific to the 60M
model.


 \begin{table*}[t]
  \centering
  \setlength{\tabcolsep}{4pt}
  \renewcommand{\arraystretch}{1.0}
  \small
  \begin{tabular}{lcccc}
  \toprule
  Dataset & Lag (Run 1/2/3) & $p$-value (Run 1/2/3) & \# Sig. \\
  \midrule
  Det.--Noun Agr.\ 1 & 500 / 200 / 100 & 0.001 / 0.003 / 0.087 & 2 \\
  Det.--Noun Agr.\ 2 & 0 / 0 / 0 & 0.682 / 0.911 / 0.706 & 0 \\
  Subj.--Verb Agr.\ 1 & 500 / 300 / 400 & 0.001 / 0.001 / 0.001 & 3 \\
  Subj.--Verb Agr.\ 2 & 300 / 900 / N/A & 0.001 / 0.001 / N/A & 2 \\
  Irr.\ Past Participle & 500 / 600 / 500 & 0.001 / 0.001 / 0.002 & 3 \\
  Exist.\ There Raising & 500 / 400 / 400 & 0.024 / 0.030 / 0.002 & 3 \\
  \bottomrule
  \end{tabular}
    \caption{Full permutation-test results across three seeds. Lag is the number of training steps between the proxy-train and proxy-validation splits reaching 80\% accuracy. The empirical p-value is computed as $(c+1)/(n+1)$, where $c$ is the number of 1000 matched random permutations with lag greater than or equal to the observed lag. The Sig. column reports the number of seeds with $p < 0.05$.}
  \label{tab:permutation_pvalues}
  \end{table*}

\section{Delayed-Generalization transition points}
\label{sec:t_after}

Table~\ref{tab:t_after} reports the delayed-generalization transition point
$t_{\text{after}}$ for each dataset. We define $t_{\text{after}}$ as
the first saved pre-training checkpoint at which accuracy on the proxy
validation split reaches 80\%. Because checkpoints are saved every 100
steps, this value is a checkpoint-level estimate of the transition time.

For each dataset, we compute $t_{\text{after}}$ separately for each of
the three independently trained 60M-parameter LLaMA seeds and report the
median value. These median transition points are used for the interpretability analyses.

\begin{table}[h]
\centering

\begin{tabular}{lc}
\toprule
Dataset & $t_{\text{after}}$ \\
\midrule
Det.--Noun Agr.\ 1            & 700  \\
Reg.\ Subj.--Verb Agr.\ 1    & 600  \\
Reg.\ Subj.--Verb Agr.\ 2    & 800  \\
Irr.\ Past Part.\ Verb        & 1000 \\
Exist.\ There Raising          & 1100 \\
\bottomrule
\end{tabular}
\caption{Dataset-specific $t_{\text{after}}$ values used in the interpretability analyses. $t_{\text{after}}$ is the first saved pre-training checkpoint at which proxy-validation accuracy reaches 80\%, computed separately for each seed and reported as the median across three 60M runs.}
\label{tab:t_after}
\end{table}

\begin{table*}[t]
\centering
\small
\resizebox{\linewidth}{!}{%
\begin{tabular}{lccccccc}
\toprule
\textbf{Dataset} & \textbf{100} & \textbf{200} & \textbf{300} & \textbf{400} & \textbf{500} & \textbf{600} & \textbf{2000} \\
\midrule
Det.-Noun 1
& -1.87 (0.21)
& -2.34 (0.17)
& -1.61 (0.28)
& -0.82 (0.44)
& +0.23 (0.70)
& +0.57 (0.72)
& +3.12 (0.90) \\

Subj.-Verb 1
& -0.05 (0.54)
& -0.26 (0.51)
& +0.21 (0.57)
& +0.60 (0.64)
& +1.16 (0.70)
& +1.89 (0.82)
& +2.63 (0.90) \\

Subj.-Verb 2
& -0.19 (0.49)
& -0.31 (0.47)
& -0.19 (0.50)
& +0.38 (0.61)
& +1.03 (0.68)
& +1.50 (0.69)
& +2.84 (0.83) \\

Irr. Past Part.
& +0.20 (0.57)
& -0.45 (0.47)
& +0.37 (0.68)
& +0.78 (0.69)
& +0.92 (0.70)
& +1.49 (0.75)
& +2.26 (0.79) \\

Exist. There
& +1.20 (0.59)
& +1.73 (0.64)
& +0.99 (0.52)
& +1.21 (0.54)
& +1.65 (0.60)
& +1.77 (0.60)
& +4.03 (0.82) \\

\midrule
\textbf{Training loss}
& \textbf{7.59}
& \textbf{6.78}
& \textbf{6.12}
& \textbf{5.59}
& \textbf{5.44}
& \textbf{4.91}
& \textbf{4.00} \\
\bottomrule
\end{tabular}%
}
\caption{Continuous log-likelihood margin and accuracy across pre-training checkpoints. Each dataset entry reports the mean log-likelihood margin, with accuracy in parentheses. Positive margins indicate higher likelihood assigned to the grammatical sentence. The final row reports C4 training loss.}
\label{tab:margin-training-loss}
\end{table*}

\section{Concept Vectors}
\label{app:concept_auroc_all_layers}

\begin{figure*}[t]
    \centering
    \includegraphics[width=\textwidth]{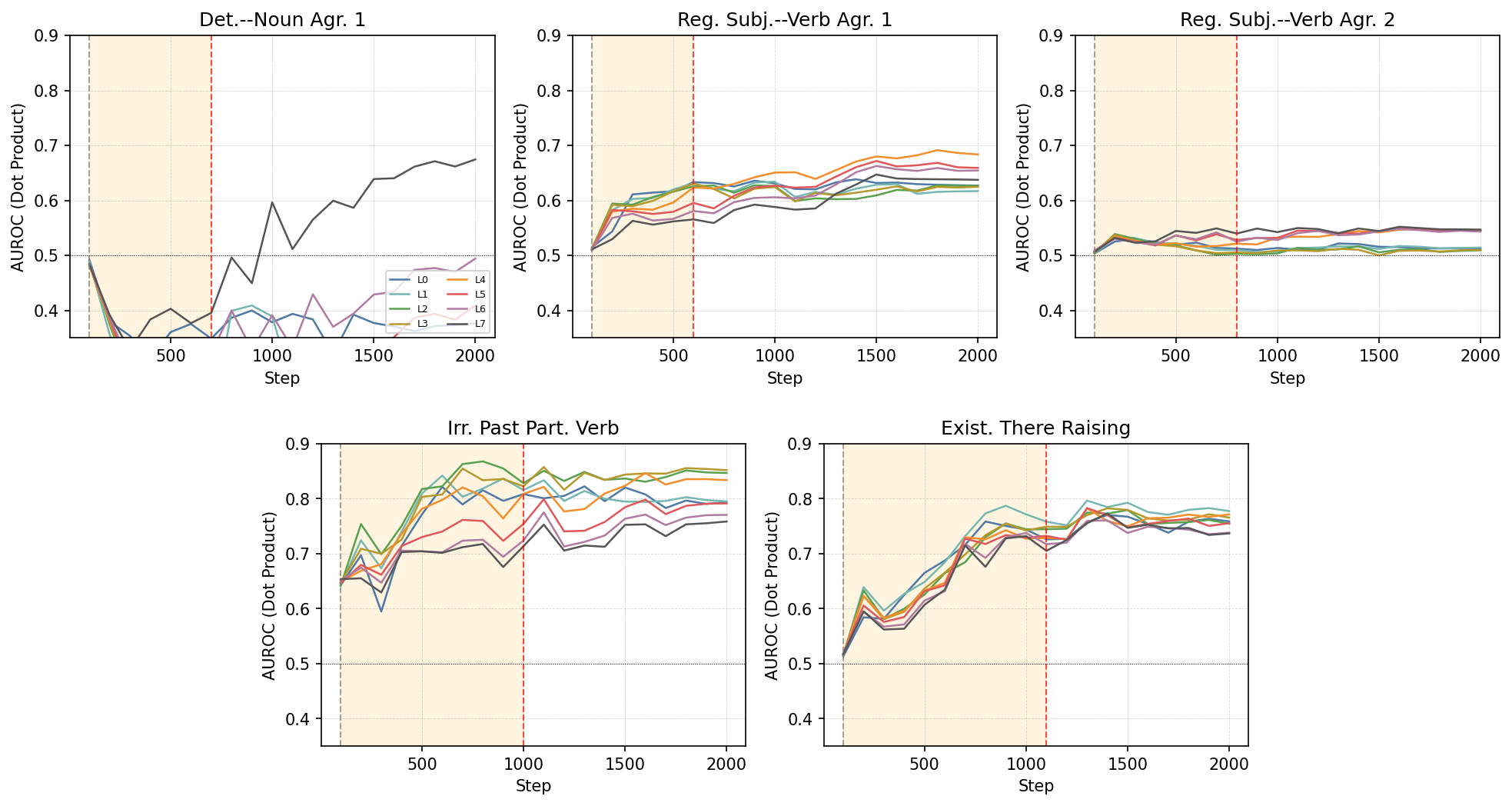}
    \caption{Concept-vector AUROC on the proxy-validation split across all layers. At each checkpoint and layer, we compute a mean-difference vector from proxy-train pairs and measure how well dot-product scores separate grammatical from ungrammatical proxy-validation sentences using last-token representations from that layer.}
    \label{fig:concept_auroc_all_layers}
\end{figure*}
Figure~\ref{fig:concept_auroc_all_layers} shows the same concept-vector AUROC analysis across all layers. The main trends from the final-layer analysis are broadly preserved, but the strength of the grammaticality signal varies by layer and phenomenon. Existential-there raising and irregular past participles show consistent separability across many layers, with AUROC increasing early and remaining well above chance. Regular subject--verb agreement shows a weaker but still positive trend: the first subject--verb dataset improves gradually across layers, while the second remains closer to chance. Determiner--noun agreement is more layer-dependent, with the clearest increase appearing in the later layers. Overall, the all-layer results support the main finding that grammaticality directions strengthen during pre-training, while also showing that the layer at which the signal is most visible depends on the grammatical phenomenon.

  \begin{figure*}[t]                                                          
      \centering                                        
      \includegraphics[width=\textwidth]{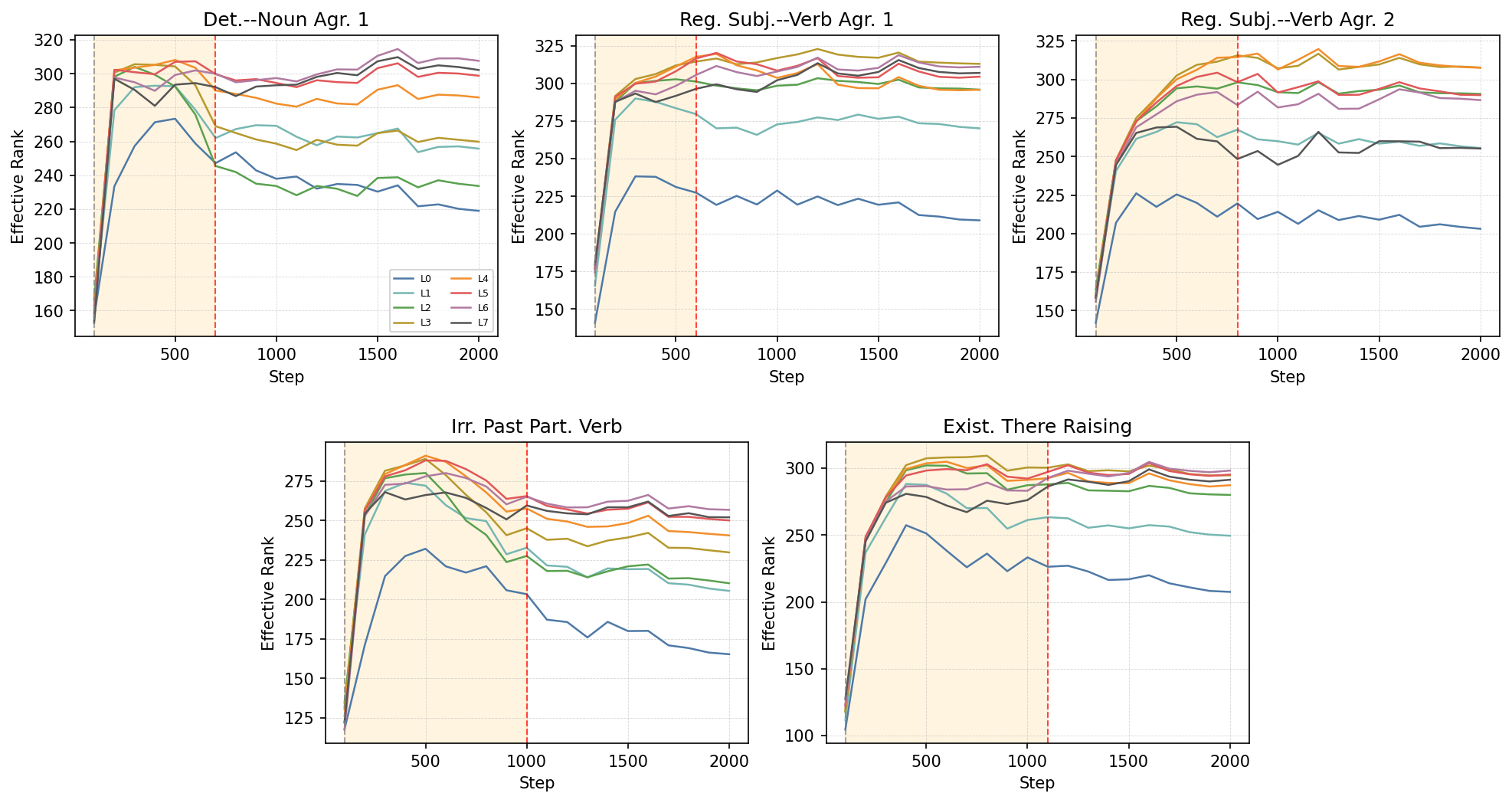}                               \caption{Effective rank of grammatical concept vectors across all layers and pre-training checkpoints. Higher effective rank indicates that the grammatical concept vectors span more independent directions. Dashed vertical lines mark $t_{\text{before}}=100$ and dataset-specific $t_{\text{after}}$; the shaded region marks the delayed-generalization window.}
      \label{fig:rank_concept_all_layers}                         
  \end{figure*} 

Figure~\ref{fig:rank_concept_all_layers} shows effective rank across all layers. The overall pattern supports the main result that grammatical concept vectors become spread across more directions after the delayed-generalization window. Across different grammatical phenomena, effective rank rises sharply early in training, indicating that the grammatical contrast quickly becomes distributed across a broader subspace. This effect is most stable in the middle and later layers, which generally maintain higher rank than the earliest layers. The dynamics around \(t_{\text{after}}\) are phenomenon-dependent: some datasets continue to increase or remain high after the delayed-generalization point, while others peak earlier and then stabilize or decline slightly.

\section{Attention}
\label{app:attention}

\begin{figure*}[t]
    \centering
    \includegraphics[width=\textwidth]{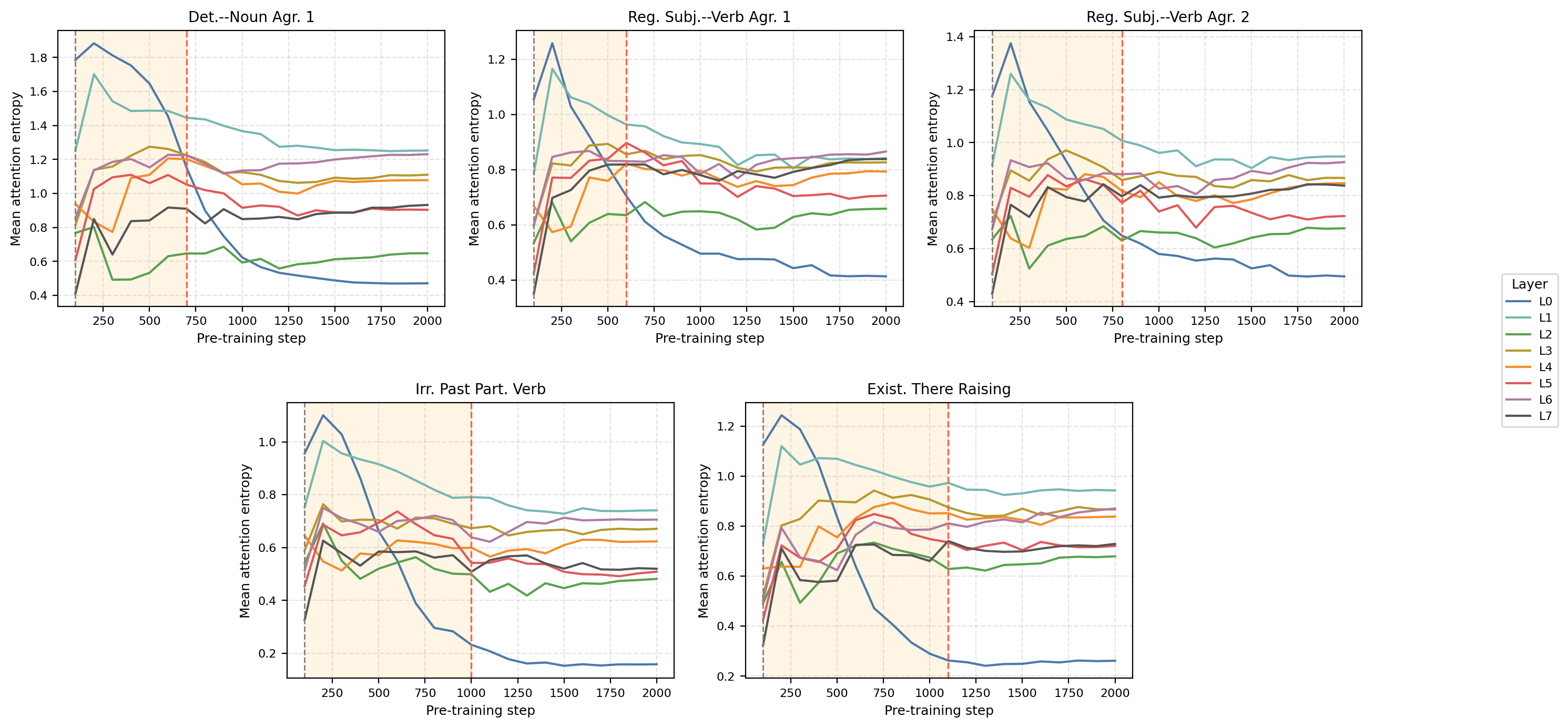}
    \caption{Mean attention entropy at the critical token across pre-training checkpoints. Each panel shows one BLiMP phenomenon; each line shows one layer, averaged over all 8 heads. Lower entropy indicates that attention from $t_{\text{critical}}$ is concentrated on fewer preceding tokens, while higher entropy indicates more diffuse attention.}
    \label{fig:attn_entropy}
\end{figure*}

\paragraph{Attention entropy.}
We measure attention entropy at \(t_{\mathrm{critical}}\) to test whether
attention becomes more concentrated after delayed generalization. For each
layer and head, we compute the entropy of the attention distribution from
\(t_{\mathrm{critical}}\) over all preceding tokens:

\[
H =
-\sum_j \alpha_j \log \alpha_j,
\]

where \(\alpha_j\) is the attention weight from \(t_{\mathrm{critical}}\) to
token \(j\). 
Lower
entropy indicates that attention is concentrated on fewer tokens, while higher
entropy indicates more diffuse attention. Figure~\ref{fig:attn_entropy}
shows that entropy often decreases or stabilizes after \(t_{\mathrm{after}}\),
consistent with attention becoming less diffuse during the same window in
which attention-to-context increases.

\begin{figure*}[t]                                            
\centering                                        
\includegraphics[width=0.9\textwidth]{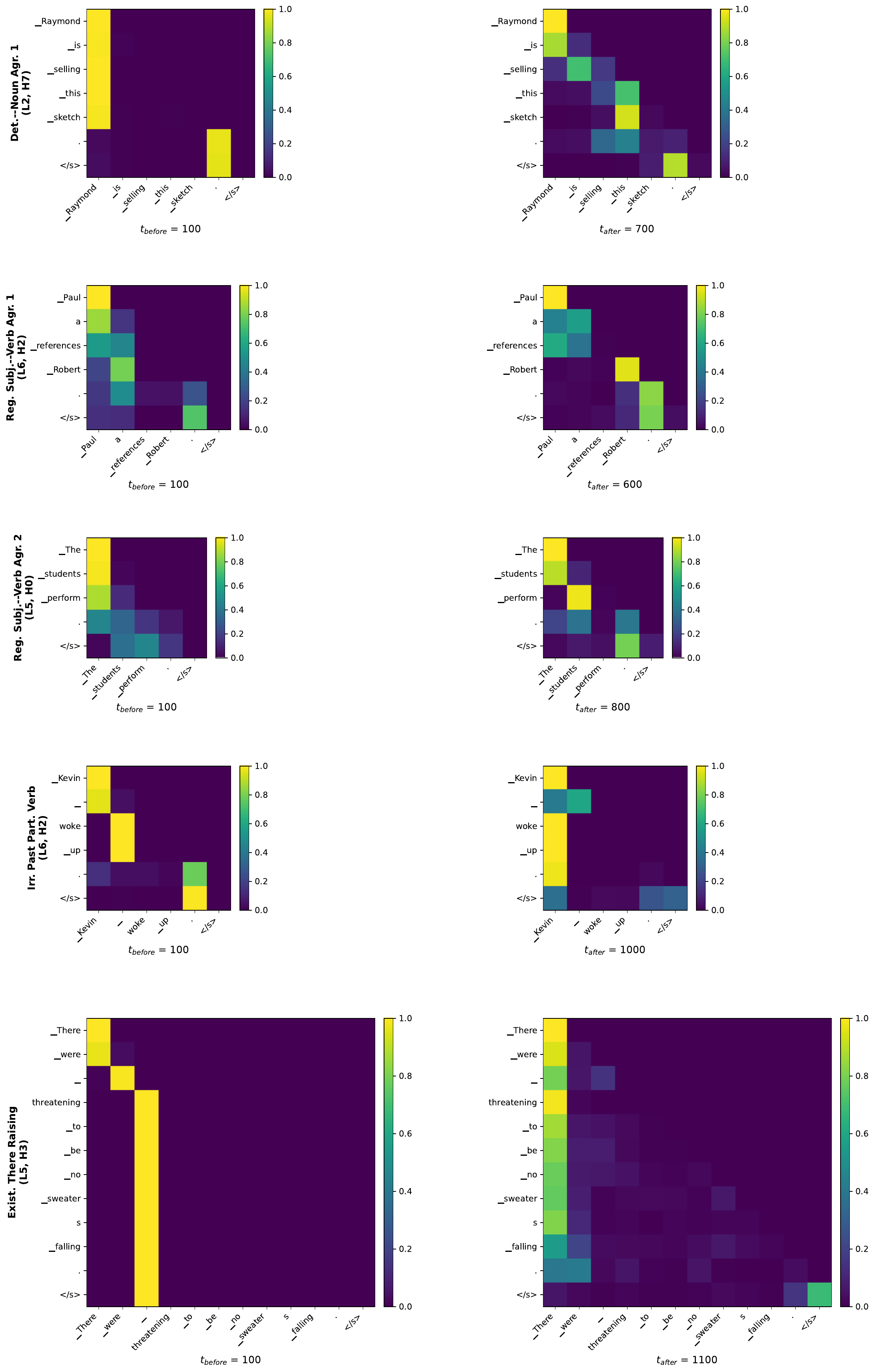}                               
\caption{Token-level attention heatmaps for the heads with the largest increase in attention-to-context after delayed generalization. Each row shows one BLiMP phenomenon and the selected layer-head pair. The left column shows attention at $t_{\text{before}}=100$, and the right column shows attention at the dataset-specific $t_{\text{after}}$. Brighter cells indicate larger attention weights.}
\label{fig:token_attn}                         
\end{figure*}  

\paragraph{Token-Level Attention Patterns.} Figure~\ref{fig:token_attn} shows token-level attention heatmaps for representative examples from the five analyzed BLiMP phenomena. Each row compares the selected head at \(t_{\text{before}}=100\) and at the corresponding \(t_{\text{after}}\). Before delayed generalization, attention is often dominated by early tokens or simple positional patterns. After \(t_{\text{after}}\), the same heads exhibit more structured attention patterns that are better aligned with the relevant grammatical dependency. Figure~\ref{fig:token_attn} shows representative token-level attention heatmaps for the best attention-to-context head in each phenomenon. For Determiner--Noun Agreement 1, the critical noun \emph{sketch} shifts from attending mostly to the sentence-initial token \emph{Raymond} at \(t_{\text{before}}\) to attending more strongly to the  determiner \emph{this} at \(t_{\text{after}}\). For Regular Subject--Verb Agreement 1, the changed verb \emph{references} increases attention to the subject \emph{Paula} after grokking. For Regular Subject--Verb Agreement 2, the verb \emph{perform} similarly attends more to the subject noun \emph{students} after grokking, matching the dependency needed to evaluate subject--verb number agreement. For Irregular Past Participle Verbs, the changed verb form \emph{woke} shows stronger attention to the preceding subject \emph{Kevin} after \(t_{\text{after}}\), consistent with noun--verb context used to distinguish valid and invalid participial forms.

\begin{table}[t]
\centering
\small
\begin{tabular}{llr}
\toprule
Dataset & Best Head & $\Delta$ Attn \\
\midrule
Det.-Noun Agr.\ 1 & L2 H7 & $+0.81$ \\
Reg. Subj.-Verb Agr.\ 1 & L6 H2 & $+0.35$ \\
Reg. Subj.-Verb Agr.\ 2 & L5 H0 & $+0.63$ \\
Irr. Past Part. Verb & L6 H2 & $+0.35$ \\
Exist. There Raising & L5 H3 & $+0.61$ \\
\bottomrule
\end{tabular}
\caption{Summary of the Head with the largest increase in attention-to-context for each dataset. $\Delta$ Attn is computed as the average attention from $t_{\text{critical}}$ to $t_{\text{context}}$ at $t_{\text{after}}$ minus the same quantity at $t_{\text{before}}$. The strongest increases occur in different layers and heads across grammatical phenomena.}
\label{tab:best_heads}
\end{table}

\begin{table*}[t]
\centering
\small
\begin{tabular}{ccrrrrr}
\toprule
Layer & Head & Det-Noun 1 & Subj-Verb 1 & Subj-Verb 2 & Irr-Past-Part & Exist-There \\
\midrule
0 & 0 & -0.07 & -0.10 & -0.12 & -0.17 & -0.18 \\
0 & 1 & -0.03 & -0.25 & -0.24 & -0.34 & -0.32 \\
0 & 2 & -0.04 & -0.03 & -0.04 & -0.20 & -0.21 \\
0 & 3 & -0.02 & -0.11 & -0.13 & -0.21 & -0.22 \\
0 & 4 & +0.00 & -0.08 & -0.09 & -0.17 & -0.19 \\
0 & 5 & -0.13 & -0.18 & -0.15 & -0.34 & -0.29 \\
0 & 6 & -0.05 & -0.18 & -0.16 & -0.28 & -0.26 \\
0 & 7 & -0.18 & -0.13 & -0.10 & -0.23 & -0.15 \\
\addlinespace
1 & 0 & +0.11 & -0.14 & -0.05 & -0.11 & -0.23 \\
1 & 1 & +0.08 & -0.19 & -0.13 & -0.31 & -0.40 \\
1 & 2 & +0.09 & -0.17 & -0.10 & -0.20 & -0.35 \\
1 & 3 & +0.26 & -0.06 & -0.10 & -0.02 & -0.03 \\
1 & 4 & -0.22 & -0.00 & -0.08 & -0.11 & -0.09 \\
1 & 5 & +0.27 & -0.06 & +0.03 & -0.18 & -0.45 \\
1 & 6 & -0.04 & -0.27 & -0.22 & -0.31 & -0.21 \\
1 & 7 & -0.04 & -0.10 & -0.11 & -0.13 & -0.02 \\
\addlinespace
2 & 0 & +0.02 & -0.09 & +0.13 & -0.06 & -0.43 \\
2 & 1 & -0.65 & -0.10 & -0.15 & -0.29 & -0.10 \\
2 & 2 & +0.68 & -0.15 & -0.06 & -0.21 & -0.43 \\
2 & 3 & -0.15 & +0.04 & +0.31 & +0.09 & -0.03 \\
2 & 4 & +0.43 & -0.04 & +0.22 & +0.29 & +0.16 \\
2 & 5 & +0.02 & -0.22 & -0.06 & -0.20 & -0.44 \\
2 & 6 & +0.29 & -0.16 & -0.01 & -0.27 & -0.35 \\
2 & 7 & \textbf{+0.77} & +0.10 & +0.41 & +0.30 & -0.10 \\
\addlinespace
3 & 0 & +0.02 & -0.24 & -0.18 & -0.33 & -0.36 \\
3 & 1 & +0.20 & -0.33 & -0.20 & -0.39 & -0.76 \\
3 & 2 & +0.47 & -0.11 & -0.01 & +0.01 & -0.16 \\
3 & 3 & +0.02 & -0.12 & +0.11 & -0.18 & +0.03 \\
3 & 4 & +0.18 & -0.08 & -0.00 & -0.11 & -0.24 \\
3 & 5 & -0.21 & -0.18 & -0.13 & -0.25 & -0.33 \\
3 & 6 & -0.49 & +0.07 & +0.08 & -0.13 & +0.43 \\
3 & 7 & +0.21 & -0.07 & -0.02 & -0.03 & -0.16 \\
\addlinespace
4 & 0 & +0.07 & -0.21 & -0.09 & -0.17 & -0.00 \\
4 & 1 & +0.25 & -0.07 & +0.24 & +0.15 & -0.35 \\
4 & 2 & +0.14 & -0.14 & +0.05 & -0.16 & -0.58 \\
4 & 3 & +0.00 & -0.06 & +0.07 & -0.07 & -0.31 \\
4 & 4 & +0.18 & -0.26 & -0.01 & -0.47 & -0.49 \\
4 & 5 & +0.03 & -0.13 & +0.01 & -0.04 & -0.11 \\
4 & 6 & +0.09 & -0.09 & -0.13 & +0.01 & +0.22 \\
4 & 7 & +0.27 & +0.16 & +0.34 & +0.26 & -0.09 \\
\addlinespace
5 & 0 & +0.21 & +0.11 & \textbf{+0.62} & +0.20 & +0.20 \\
5 & 1 & +0.02 & -0.19 & +0.04 & -0.31 & -0.50 \\
5 & 2 & -0.25 & +0.17 & +0.38 & -0.18 & -0.47 \\
5 & 3 & +0.14 & -0.08 & -0.12 & -0.02 & \textbf{+0.61} \\
5 & 4 & -0.03 & -0.14 & -0.05 & -0.30 & -0.27 \\
5 & 5 & +0.08 & -0.30 & -0.24 & -0.29 & +0.02 \\
5 & 6 & +0.55 & +0.03 & +0.08 & +0.00 & +0.09 \\
5 & 7 & -0.12 & -0.01 & -0.06 & -0.14 & +0.10 \\
\addlinespace
6 & 0 & +0.09 & -0.07 & +0.32 & +0.14 & -0.06 \\
6 & 1 & +0.05 & -0.03 & +0.25 & +0.09 & +0.09 \\
6 & 2 & +0.14 & \textbf{+0.35} & +0.48 & \textbf{+0.35} & +0.20 \\
6 & 3 & +0.01 & -0.30 & -0.18 & -0.24 & -0.47 \\
6 & 4 & +0.07 & +0.16 & +0.19 & +0.04 & -0.03 \\
6 & 5 & +0.15 & -0.09 & +0.06 & -0.33 & -0.25 \\
6 & 6 & +0.06 & +0.18 & +0.27 & +0.15 & -0.35 \\
6 & 7 & +0.50 & +0.10 & +0.12 & +0.13 & +0.15 \\
\addlinespace
7 & 0 & +0.03 & +0.12 & +0.52 & +0.18 & -0.07 \\
7 & 1 & +0.12 & -0.23 & +0.00 & -0.22 & -0.50 \\
7 & 2 & +0.10 & -0.35 & -0.28 & -0.11 & +0.01 \\
7 & 3 & -0.34 & +0.07 & +0.05 & +0.05 & +0.01 \\
7 & 4 & -0.25 & -0.24 & -0.12 & -0.47 & -0.02 \\
7 & 5 & -0.13 & -0.07 & -0.05 & -0.42 & -0.09 \\
7 & 6 & +0.13 & -0.03 & -0.08 & -0.10 & -0.03 \\
7 & 7 & -0.44 & -0.02 & +0.05 & +0.11 & +0.13 \\
\bottomrule
\end{tabular}
\caption{Change in attention-to-context from $t_{\text{before}}$ to $t_{\text{after}}$ for all layer-head combinations across the five analyzed datasets. Each entry reports $\Delta$ attention-to-context, computed as attention from $t_{\text{critical}}$ to $t_{\text{context}}$ at $t_{\text{after}}$ minus the same quantity at $t_{\text{before}}$. Bold indicates the head with the largest positive increase for each dataset.}
\label{tab:delta_attn}
\end{table*}



\end{document}